%% file: main.tex
\documentclass{article}
\usepackage{iclr2027_conference,times}

\usepackage{amsmath}
\usepackage{amssymb}
\usepackage{booktabs}
\usepackage{graphicx}
\usepackage{multirow}
\usepackage{tabularx}
\usepackage{array}
\usepackage{xcolor}
\usepackage{url}
\usepackage[hypertexnames=false]{hyperref}
\usepackage{float}
\usepackage[section]{placeins}
\usepackage[most]{tcolorbox}
\usepackage{fvextra}
\usepackage{makecell}
\usepackage{caption}
\title{Holo-World: Unified Camera, Object and Weather Control for Video World Model}

\author{Xiangchen Yin$^{1,3}$, 
Wenzhang Sun$^{2}$,
Jiahui Yuan$^{1}$,
Zijie Liu$^{1}$, \\
\textbf{Yinda Chen$^{1}$, 
Wei Li$^{1}$, 
Dachun Kai$^{1}$, 
Chunfeng Wang$^{2}$, 
Xiaoyan Sun$^{1,3,{\dag}}$} \\
$^{1}$University of Science and Technology of China \quad
$^{2}$Li Auto. \\
$^{3}$Institute of Artificial Intelligence, Hefei Comprehensive National Science Center 
}

\footnotetext{Corresponding authors. Email: \texttt{yinxc2000@gmail.com}}

\newcommand{\cmark}{\checkmark}

\newtcolorbox{promptbox}[1]{%
  enhanced,
  breakable,
  colback=blue!2,
  colframe=blue!45!black,
  coltitle=black,
  title=\textbf{#1},
  fonttitle=\small,
  boxrule=0.5pt,
  arc=1mm,
  left=1mm,
  right=1mm,
  top=1mm,
  bottom=1mm
}
\DefineVerbatimEnvironment{promptverbatim}{Verbatim}{breaklines=true,breakanywhere=true,fontsize=\scriptsize}

\AtBeginDocument{%
  \setlength{\abovedisplayskip}{2pt plus 1pt minus 1pt}%
  \setlength{\belowdisplayskip}{2pt plus 1pt minus 1pt}%
  \setlength{\abovedisplayshortskip}{1pt plus 1pt minus 1pt}%
  \setlength{\belowdisplayshortskip}{1pt plus 1pt minus 1pt}%
}
\makeatletter
\def\section{\@startsection {section}{1}{\z@}{-1.4ex plus
    -0.4ex minus -.2ex}{0.9ex plus 0.2ex
minus0.2ex}{\large\sc\raggedright}}
\def\subsection{\@startsection{subsection}{2}{\z@}{-1.2ex plus
-0.4ex minus -.2ex}{0.5ex plus .2ex}{\normalsize\sc\raggedright}}
\makeatother

\iclrfinalcopy

\begin{document}

\maketitle

\begin{figure}[!htbp]
  \centering
  \vspace{-1.2em}
  \includegraphics[width=\linewidth]{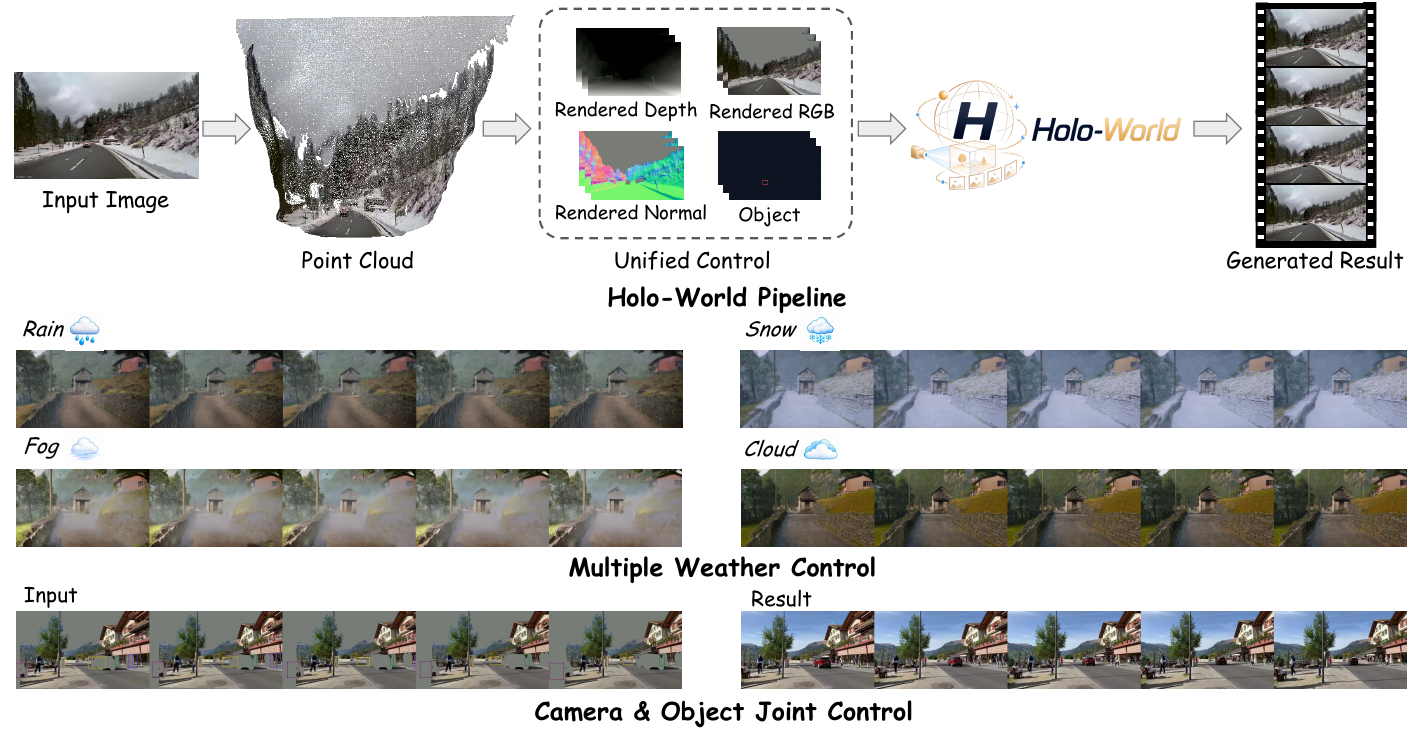}
  \vspace{-1.2em}
  \caption{\textbf{Unified state control in Holo-World.} Holo-World jointly controls camera motion, object dynamics, and weather state within the same observed world. 
  }
  \label{fig-unified-control}
\end{figure}

\begin{abstract}
\input{sections/0_abstract}

\end{abstract}

\FloatBarrier
\clearpage

\input{sections/1_introduction}

\input{sections/2_related_work}
\input{sections/3_holostatedata}

\input{sections/4_method}
\input{sections/5_experiments}
\input{sections/6_conclusion}

\subsection*{AI use statement}

We used generative video models to construct paired weather-transfer training data, Qwen3-VL to generate scene and weather annotations, and a GPT model to evaluate weather alignment and video quality. In the period of paper writing, we only exploit the Large Language Model (LLM), i.e., GPT-6, for word polishing to improve the presentation quality. No LLMs contributes to the core scientific methods, technical contents, or any substantive research contributions.



\bibliography{reference}
\bibliographystyle{iclr2027_conference}

\FloatBarrier
\clearpage
\appendix
\input{sections/A_appendix}

\end{document}

%% file: sections/0_abstract.tex

Video world models are moving toward preserving an observed world under controllable camera and object motion while allowing its environmental state to change. Yet these controls remain isolated, and weather generation typically relies on a source video or reconstructed scene that already specifies future structure. We study a first-frame-anchored source-to-state setting, where the model starts from a single image and follows explicit camera and object controls and an optional weather instruction, then generates a video that either preserves the source world or transfers it to a target weather state. To address these challenges, we first build \textbf{HoloStateData}, a state video dataset that turns diverse videos into unified control samples for camera, object, and weather supervision. Second, we introduce \textbf{Holo-World}, a unified controllable video world model that jointly controls the scene from a single image. Its Unified Scene Adapter factorizes world preservation and weather transfer into distinct parameter subspaces, using rendered background, geometry buffers, and object controls to maintain controlled scene structure while modeling weather-dependent appearance and particle effects. Additionally, Scene-Weather Decomposed CFG guides scene and weather residuals separately, strengthening target weather effects without over-amplifying the full condition. Quantitative and qualitative experiments demonstrate that Holo-World maintains precise camera and object controls with consistent scene structure while transferring scenes into diverse target weather states, outperforming video-to-video weather editing baselines on weather-state generation. Our project page is available at {\color{blue}\url{https://xiangchenyin.github.io/Holo-World/}}.


%

%% file: sections/1_introduction.tex
\section{Introduction}
\label{sec:introduction}



Video world models~\citep{genie,hunyuangamecraft,matrixgame,matrixgame2,voyager,longvie} are moving from generating visually plausible videos toward simulating dynamic worlds under structured conditions. Future observations in video world modeling depend not only on how the camera moves and how objects evolve, but also on the environmental state in which the same scene is rendered. Current controllable video methods usually model these factors separately, with camera-control methods~\citep{gen3c,neoverse,voyager} specifying viewpoint changes, object-control methods~\citep{magicmotion,wan_move,traceanything,bar2025navigation} steering dynamic entities, and weather-oriented methods~\citep{autoweather4d,weatherweaver} synthesizing or editing environmental appearance. Existing weather generation methods are largely formulated as video-to-video editing or simulation~\citep{cosmos,weatheredit,3d_gs}, where a complete source video or reconstructed scene already provides future layout, motion, and temporal continuity.

However, for a video world model these future states should be generated by the model, making joint camera, object, and weather control a different problem from video-to-video weather editing. Although recent works have begun to couple camera and object control~\citep{versecrafter,uni3c,symphomotion}, weather state is still rarely integrated into the same generation process. We therefore study a first-frame source-to-state setting, where a model observes a single image and synthesizes future frames under explicit camera and object controls while rendering the same scene into a target weather state~\citep{weatherweaver,revideo,scenecrafter}. This gap stems from two coupled obstacles: (1) the lack of data that jointly captures camera, object, and weather controls, and (2) the modeling conflict between world preservation and weather-state transfer within the same controllable model.

The data obstacle follows from the separation of existing controllable video methods. Existing methods~\citep{epic,motionctrl,weatheredit} supervise parts of this control space, but their data assumptions are not organized around a shared control interface. These advances~\citep{spatialvid,sekai} provide important control signals, but they do not supply training samples in which camera pose~\citep{cameractrl,realcami2v}, background geometry~\citep{worldmirror,vggt}, object controls~\citep{motionpro,motionbooth}, and target weather-state supervision are jointly observed. Appendix Table~\ref{tab-capability-comparison} summarizes the resulting capability gap.

The modeling obstacle is to support both background preservation and weather transfer within one model. Real videos provide supervision for source appearance and geometry consistency, while paired weather videos provide supervision for weather-state transfer under matched structural controls. If these two objectives are entangled in the same adapter parameter space, the learned residuals can either suppress weather changes or disturb camera, object, and background consistency. The same entanglement appears during sampling, where vanilla classifier-free guidance~\citep{cfg} scales a global residual that couples scene preservation with weather changes, so stronger guidance may enhance weather effects while also over-amplifying the source scene.

To provide supervision for unified control, we build \textbf{HoloStateData}, a state video dataset that organizes real videos, paired simulated weather videos, and weather-transferred videos into samples aligned with camera poses, geometry anchors, object controls, scene text, and weather-state supervision. We introduce \textbf{Holo-World}, a unified controllable video world model that jointly controls camera motion, object dynamics, and scene weather state from a single image. As shown in Figure~\ref{fig-unified-control}, Holo-World supports both background-consistent generation and weather-state transfer under the same structured control setting. To address these challenges, we introduce Unified Scene Adapter (\textbf{UniSA}), instantiated with a World Adapter and a State Adapter that share one frozen video backbone but learn separate residual subspaces for world preservation and weather transfer. UniSA grounds camera-conditioned generation with rendered background controls, uses G-buffers to stabilize scene structure, and incorporates object controls to constrain dynamic entities, allowing weather-dependent appearance and particle effects to emerge within the same controlled scene rather than relying on a cascaded video-to-video editor. During sampling, Scene-Weather Decomposed CFG (\textbf{SW-CFG}) applies the same separation at the guidance level, using a scene residual to preserve the controlled world and an independent weather residual to strengthen weather effects without over-amplifying the full condition. Extensive quantitative and qualitative experiments on our benchmark demonstrate that Holo-World maintains precise camera and object control with consistent scene structure while transferring scenes into diverse target weather states.

Our contributions are summarized as follows:
\begin{itemize}
    \item We formulate unified state control for video world models, where a single image-to-video model follows camera and object controls while supporting both world preservation and weather-state transfer.
  
    \item We build HoloStateData, a state video dataset that turns diverse videos into unified control samples for camera, object, and weather supervision.
      
    \item We introduce Holo-World with Unified Scene Adapter, which factorizes world preservation and weather-state transfer into distinct parameter subspaces within one frozen video backbone, together with Scene-Weather Decomposed CFG for independently guiding scene and weather residuals during sampling.
  
    \item Extensive experiments demonstrate that Holo-World maintains precise camera and object controls and consistent scene structure while generating scenes into diverse weather states, outperforming video-to-video weather editing baselines on weather generation.
\end{itemize}

%% file: sections/2_related_work.tex
\section{Related Work}
\label{sec:related-work}

\vspace{-0.1in}
\paragraph{Video world models.}
Video world models learn dynamic scene evolution by predicting future world states. Recent generative world models scale this paradigm with transformer and diffusion backbones for action-conditioned generation~\citep{genie,matrixgame,matrixgame2,hunyuangamecraft,worldplay}. Geometry-aware methods further incorporate explicit 3D reconstruction~\citep{worldmirror,vggt} to support world exploration and navigable scene generation~\citep{voyager,wonderworld,dimensionx,hunyuanworld2,lyra2,nerf}. These methods advance dynamic world synthesis, but their controllable state spaces are usually limited to actions, text, camera motion, or reconstructed geometry. A unified video world model interface that jointly controls camera motion, object dynamics, and environmental state remains largely unexplored.

\vspace{-0.1in}
\paragraph{Video generation with camera and object motion control.}
Controllable video generation has moved from text-only conditioning~\citep{hunyuanvideo} toward explicit structural controls. Camera-aware and 3D-aware methods~\citep{motionctrl,cameractrl,viewcrafter,gen3c,motionbooth,epic,motioni2v,realcami2v} inject trajectories, poses, rendered geometry, or learned camera motion into video diffusion models, enabling viewpoint control and image-conditioned novel-view generation. Object-control and camera-object methods~\citep{versecrafter,neoverse,uni3c,symphomotion} further steer dynamic entities with 4D geometric controls, trajectories ~\citep{3dtrajmaster,wan_move}, regions, boxes ~\citep{magicmotion,motionpro}, or latent motion guidance.

\vspace{-0.1in}
\paragraph{Weather editing and environment-state generation.}
Weather-oriented methods~\citep{weatherweaver, autoweather4d, weatheredit} synthesize or edit rain, snow, fog, wet surfaces, snow coverage, overcast skies, and related environmental effects. Some methods formulate weather control as video-to-video editing~\citep{cosmos, vace}, where a full source video already provides future layout, motion, and temporal continuity. Others rely on reconstructed 3D scenes, driving assets, multi-view inputs, maps, or multimodal video controls to constrain the target sequence. These designs are effective for weather transfer under strong evidence, but they do not directly address single-image video generation where the model must follow explicit camera and object controls while synthesizing the target weather state.

%% file: sections/3_holostatedata.tex
\section{HoloStateData}
\label{sec:dataset}

\textbf{HoloStateData} is designed around this requirement by representing each clip as a source-to-state record rather than a standalone captioned video. The first frame and source-side controls define the observed world, including camera-conditioned background motion, static geometry anchors, and object boxes. When paired weather supervision is available, the target video and weather prompt specify the scene state to synthesize. This shared interface supports both no-edit background preservation on real videos and weather-state transfer on paired weather videos.

\begin{figure}[t]
  \centering
  \includegraphics[width=\linewidth]{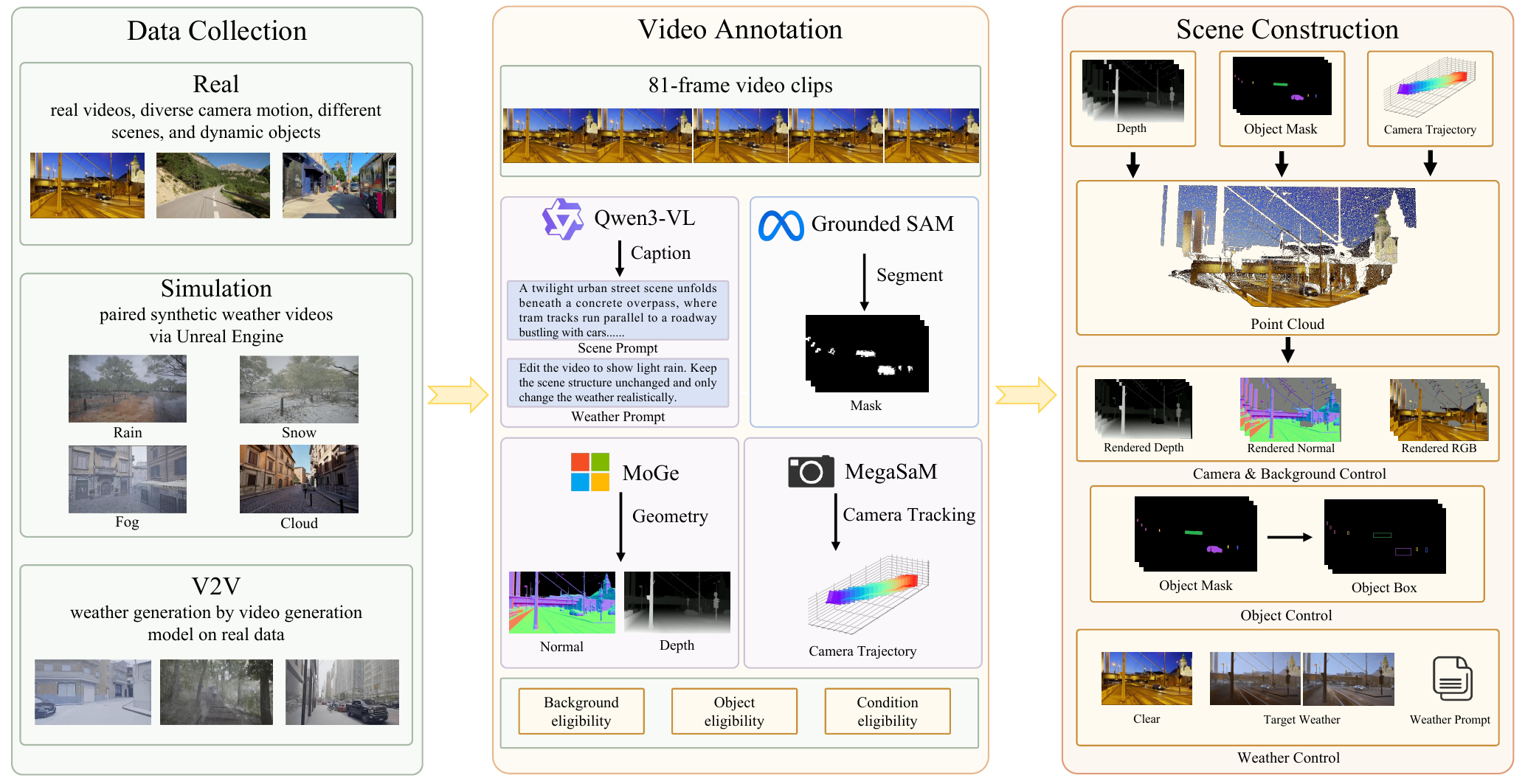}
  \vspace{-1.2em}
  \caption{\textbf{HoloStateData construction pipeline.} These annotations are converted into source-side rendered controls and object controls, while paired target-weather videos provide supervision for weather-state transfer.}
  \label{fig-dataset-pipeline}
\end{figure}

As shown in Figure~\ref{fig-dataset-pipeline}, HoloStateData is built through three stages: data collection, video annotation, and scene construction. Data collection gathers real videos, paired simulated weather videos, and weather-transferred videos via V2V generation. Video annotation extracts factorized scene and weather text, object masks, camera motion, and dense geometry. Scene construction renders source-side background RGB, depth, and normal controls, converts object masks into object controls, and associates paired samples with target-weather supervision. The central rule is that target-weather pixels are never used to render model-facing controls, they only provide the supervised target video.

\vspace{-0.1in}
\paragraph{Data sources.}
We first use SpatialVID-HQ~\citep{spatialvid} as the real-video pool, which provides in-the-wild scenes with diverse camera trajectories, object dynamics, and layouts. We then use open-source paired simulated weather videos rendered with Unreal Engine, where the same synthetic scenes are captured under clear and target weather states~\citep{weatherweaver}. These paired videos share scene structure while differing in weather appearance, providing clean supervision for weather-state transfer. Finally, we use SpatialVID-HQ clips as source videos and synthesize paired target-weather videos with a combination of our weather-transfer model and proprietary video editing models, extending weather supervision to real camera trajectories and object dynamics while complementing the more controlled synthetic pairs.

\vspace{-0.1in}
\paragraph{Automatic annotation and scene construction.}
Following in~\citep{versecrafter,spatialvid}, we generate factorized text, object, and geometry controls for each retained 81-frame clip. Qwen3-VL~\citep{qwen3vl} produces a scene prompt that describes scene identity, objects, actions, temporal evolution, and camera motion while excluding weather. Simulation and V2V samples additionally receive a weather prompt that only describes the target weather state, whereas Real samples use no weather-edit prompt. Object controls are built by detecting first-frame objects, initializing and propagating masks with Grounded-SAM2~\citep{groundingdino,sam2}, and converting each per-frame mask into a tight bounding box as model-facing object control. For camera/static-world control, Depth Anything~\citep{depthanythingv2} and UniDepth~\citep{unidepth} provide monocular depth priors, MegaSaM~\citep{droid_slam,megasam} estimates camera parameters, and MoGe~\citep{moge} supplies dense depth and normal anchors for rendering source-side background RGB, depth, and normal videos. Scene construction enforces source-target decoupling: target videos and weather prompts supervise the desired weather state, but rendered background controls and object boxes are always built from the source side. This keeps world-consistent controls separate from weather-state supervision. Appendix~\ref{app:dataset-snapshot} provides additional details on our dataset.

\paragraph{Data subsets.}
These sources are organized into three subsets with complementary supervision roles. The Real subset stabilizes the source-world interface using real camera motion, object dynamics, and scene layouts without a weather-edit target. The Simulation subset provides aligned synthetic weather pairs with shared scene structure and controlled target-weather changes. The V2V subset exposes the model to weather transfer on real videos, complementing the synthetic pairs with real trajectories and dynamic entities. After filtering, HoloStateData contains 15K training samples. We also reserve a mutually separated 150-sample benchmark dataset, organized into Real, Simulation, and V2V subsets with 50 samples each.

\vspace{-0.1in}
\paragraph{Metrics.}
On the Real subset, we report VBench-I2V~\citep{vbench++} for video quality, rotation error (RotErr), translation error (TransErr)~\citep{cameractrl} and ObjMC~\citep{motionctrl} for camera and object-control errors, and background metrics for background preservation. The background metrics compute PSNR, SSIM, and LPIPS between generated frames and rendered RGB on valid static regions, so they directly measure whether the controlled background remains stable. Weather metrics are not reported on the Real subset because there is no target weather state. On the Weather subset (Simulation/V2V), we report Weather Alignment, VLM Evaluation, and User Study. Weather Alignment uses the target weather prompt and generated video to judge whether the requested core weather condition appears in the result, and reports the aligned-sample ratio. VLM Evaluation uses the same target weather prompt and generated video, but asks a VLM evaluator to score weather background editing, particle dynamics, atmospheric coherence, and overall visual quality on a 0--100 score. User Study reports blind pairwise human preference between Holo-World and each weather-editing baseline. Appendix~\ref{app:metrics} gives the weather evaluation prompts and aggregation rules, and Appendix~\ref{app:background-preserve} details the background preservation metrics.

\begin{figure}[t]
  \centering
  \includegraphics[width=\linewidth]{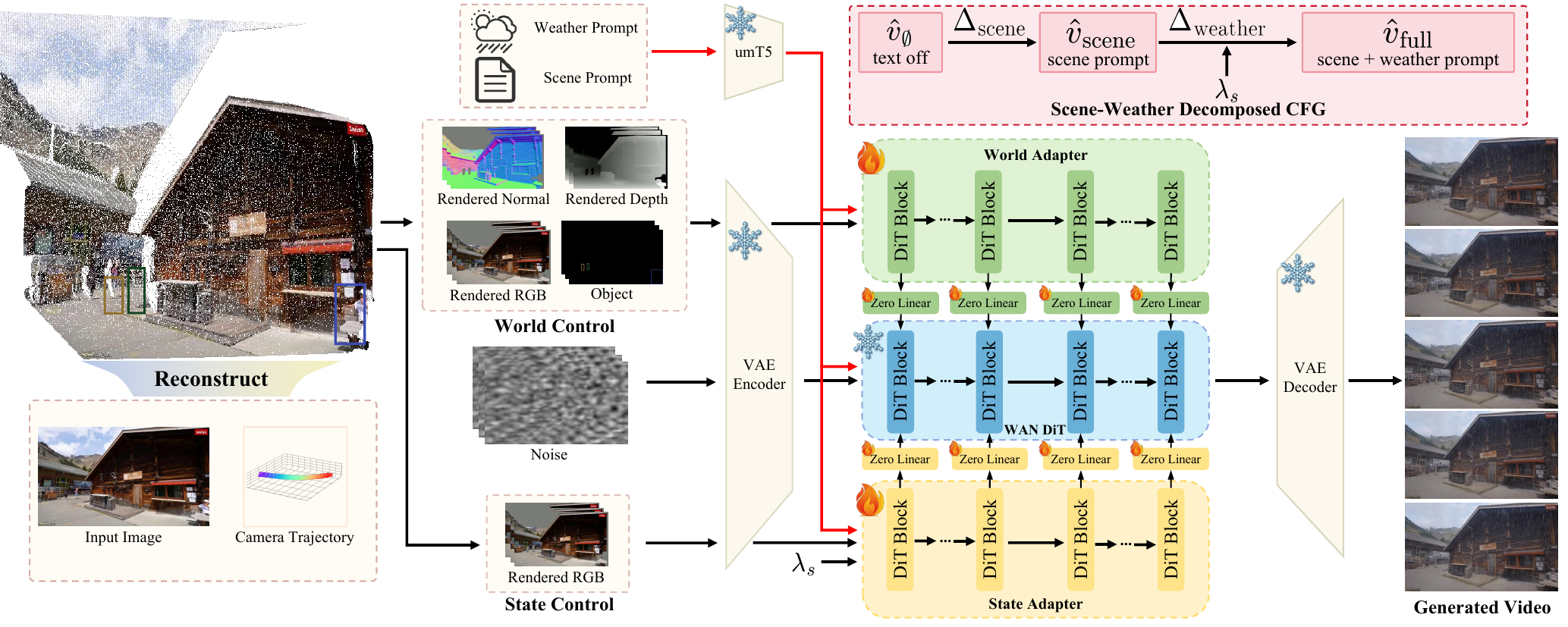}
  \vspace{-1.2em}
  \caption{\textbf{Overview of Holo-World.} Given a first frame and factorized source-to-state controls, Holo-World decomposes controllable video generation into world-preservation and weather-transfer residual paths.}  \label{fig-method}
\end{figure}

%% file: sections/4_method.tex
\vspace{-0.1in}
\section{Holo-World}
\label{sec:method}

Holo-World is built around a factorized source-to-state interface that contains the observed world to preserve and the target weather state to transfer. Unified Scene Adapter (UniSA) injects structured controls into a frozen Wan base model through preservation-oriented and weather-oriented residual subspaces, while Scene-Weather Decomposed CFG (SW-CFG) applies the same separation at inference time to avoid a single global guidance direction that entangles scene preservation and weather transfer. Figure~\ref{fig-method} summarizes the architecture.

\subsection{Control Interface and Training}
\label{subsec:problem}

Holo-World uses a first-frame source-to-state interface with structured controls and factorized text. Given the input frame $I_0$, $C_{\mathrm{world}}$ represents source-side world controls, $C_{\mathrm{state}}$ is the source appearance anchor for state transfer, $c_{\mathrm{scene}}$ is the scene prompt, and $c_{\mathrm{state}}$ is the weather prompt. All rendered controls are constructed from the source side before generation. During training, the automatic construction pipeline uses the first frame, camera trajectory, object masks, and dense source geometry to render background controls and derive object boxes. At inference, the same interface can be filled by an explicit camera trajectory, object controls from Grounded-SAM2, or user interaction, and dense first-frame geometry from MoGe. Target-weather videos are never used to render these controls, so the model does not receive target-side appearance leakage. The world and state controls are
\begin{equation}
C_{\mathrm{world}} = \{R_{\mathrm{rgb}},R_{\mathrm{depth}},R_{\mathrm{normal}},C_{\mathrm{bbox}}\},
\qquad
C_{\mathrm{state}} = \{R_{\mathrm{rgb}}\}.
\label{eq:control-groups}
\end{equation}
where $R_{\mathrm{rgb}}$, $R_{\mathrm{depth}}$, and $R_{\mathrm{normal}}$ are RGB, depth, and normal videos rendered from the first-frame geometry and camera trajectory, and $C_{\mathrm{bbox}}$ is the object control video derived from propagated masks. The same rendered RGB is also used as $C_{\mathrm{state}}$, serving as a source appearance anchor for weather-state transformation rather than a target-weather image or generated weather proxy.

Rendered background controls specify camera-conditioned static-world evolution, bbox videos constrain dynamic entities, and the weather prompt defines the target scene state relative to the source appearance anchor. Depth and normal maps serve as geometric anchors of structure consistency, reducing floating texture artifacts and structure drift.

Holo-World keeps the Wan flow-matching~\citep{flow_matching} objective and only updates UniSA parameters. Let $c=\operatorname{merge}(c_{\mathrm{scene}},c_{\mathrm{state}})$ be the global text condition received by the frozen backbone. Given the noisy latent $z_t$ at timestep $t$, the model predicts the Wan flow target $v_t$:
\begin{equation}
\mathcal{L}_{\mathrm{fm}}
=
\mathbb{E}_{\mathcal{D},\epsilon,t}
\left[
\left\|
f_{\theta,\phi}(z_t,t,I_0,C_{\mathrm{world}},C_{\mathrm{state}},c) - v_t
\right\|_2^2
\right].
\label{eq:training-objective}
\end{equation}
The VAE, text encoder, and Wan backbone parameters $\theta$ are frozen, and $\phi=\{\phi_w,\phi_s\}$ represent the trainable World and State Adapter parameters. Real samples supervise no-edit preservation with $V_{\mathrm{origin}}$, while Simulation and V2V samples supervise weather-state transfer with $V_{\mathrm{target}}$.

\subsection{Unified Scene Adapter}
\label{subsec:unisa}

The central modeling conflict is that background preservation and weather-state generation require different residual behaviors. Real samples supervise source appearance, camera-consistent background structure, and object-control adherence. Weather-paired samples keep the same structural controls but require changes in visibility, sky condition, wetness, snow coverage, fog density, and other scene-state cues. If both objectives are absorbed by the same adapter parameter space, preservation supervision can suppress weather changes, while weather supervision can disturb geometry and object adherence.

Unified Scene Adapter addresses this conflict with one adapter interface and two parameter-disjoint subspaces. The World Adapter learns preservation residuals for camera, background geometry, and object layout, and the State Adapter learns weather-transfer residuals from the source appearance anchor and target weather text:
\begin{align}
h^w_{\ell} &= A^w_{\ell}(C_{\mathrm{world}},c_{\mathrm{scene}}),\\
h^s_{\ell} &= A^s_{\ell}(C_{\mathrm{state}},c_{\mathrm{state}}).
\label{eq:unisa-hints}
\end{align}
where $A^w_{\ell}$ and $A^s_{\ell}$ are the World and State Adapter blocks aligned with the $\ell$-th injected Wan block. The two blocks have the same DiT structure, produce residual hints with the same hidden shape, and do not share trainable parameters. They operate on the same frozen video backbone and therefore form a single end-to-end model rather than a cascaded weather editor.

Let $F_\ell$ represents the $\ell$-th frozen Wan DiT block and let $x_{\ell-1}$ represents the input hidden state. For injected layers $\ell\in\mathcal{L}$, UniSA augments the backbone update with the two residual hints:
\begin{equation}
x_\ell =
F_\ell(x_{\ell-1},c)
+ \lambda_w h^w_\ell
+ \lambda_s h^s_\ell .
\label{eq:unisa-update}
\end{equation}
Layers outside $\mathcal{L}$ follow the original pretrained Wan forward process. We use $\mathcal{L}=\{0,5,10,15,20,25\}$, initialize the adapter blocks from their corresponding Wan blocks. The world scale is fixed as $\lambda_w=1$, while the state scale follows the sample type:
\begin{equation}
\lambda_s =
\begin{cases}
0, & \text{without weather prompt},\\
1, & \text{with weather prompt}.
\end{cases}
\label{eq:state-scale}
\end{equation}
This gate keeps Real samples from forcing the State Adapter to learn identity appearance changes, while Simulation and V2V samples train state transformation on top of the same source controls.

\subsection{Scene-Weather Decomposed CFG}
\label{subsec:cfg}

Even though UniSA separates world and state residuals inside the network, vanilla classifier-free guidance still extrapolates along one empty-to-full direction, jointly strengthening scene structure, weather cues, appearance shifts, and text interactions, which can enhance weather effects at the cost of source-world preservation:
\begin{equation}
\hat{v}
=
\hat{v}_{\emptyset} + s(\hat{v}_{\mathrm{full}}-\hat{v}_{\emptyset}).
\end{equation}
where $\hat{v}_{\emptyset}$ is the empty inference, $\hat{v}_{\mathrm{full}}$ is the full inference, and $s$ is the guidance scale. In Holo-World, the direction $\hat{v}_{\mathrm{full}}-\hat{v}_{\emptyset}$ is not a pure weather direction. It mixes scene semantics, weather semantics, color and texture shifts, World Adapter effects, State Adapter effects, and control-text interactions. Increasing scale $s$ can make weather more visible, but the same coefficient may over-drive no-edit Real samples with color shifts, over-saturation, or texture artifacts. Reducing scale $s$ preserves Real samples, but weakens weather generation.

Scene-Weather Decomposed CFG follows the UniSA factorization at sampling time. At each denoising step, all branches share the same latent $z_t$, controls, timestep, and model parameters, only the scene and weather prompt change. The sampler evaluates three predictions:
\begin{align}
\hat{v}_{\emptyset} &=
f_\theta(z_t,t,I_0,C_{\mathrm{world}},C_{\mathrm{state}},c_{\emptyset}),\\
\hat{v}_{\mathrm{scene}} &=
f_\theta(z_t,t,I_0,C_{\mathrm{world}},C_{\mathrm{state}},c_{\mathrm{scene}}),\\
\hat{v}_{\mathrm{full}} &=
f_\theta(z_t,t,I_0,C_{\mathrm{world}},C_{\mathrm{state}},c_{\mathrm{scene}},c_{\mathrm{state}}).
\label{eq:cfg-branches}
\end{align}
The three predictions define two residual directions. The empty prediction serves as a control-only baseline with text removed. The scene prediction restores the scene prompt and measures the residual needed to preserve the controlled world relative to this baseline. The full prediction further includes target-weather text, making $\hat{v}_{\mathrm{full}}-\hat{v}_{\mathrm{scene}}$ the weather-specific residual. For Real samples, no target weather is requested, so $\lambda_s=0$ removes state residual. For Simulation and V2V samples, $\lambda_s=1$ keeps it active for weather-state transfer.

The guidance direction is then decomposed into a scene residual and a weather residual:
\begin{equation}
\Delta_{\mathrm{scene}} = \hat{v}_{\mathrm{scene}}-\hat{v}_{\emptyset},
\qquad
\Delta_{\mathrm{weather}} = \hat{v}_{\mathrm{full}}-\hat{v}_{\mathrm{scene}}.
\end{equation}
\begin{equation}
\hat{v}
=
\hat{v}_{\emptyset}
+ s_{\mathrm{scene}}\Delta_{\mathrm{scene}}
+ \lambda_s s_{\mathrm{weather}}\Delta_{\mathrm{weather}}.
\label{eq:scene-weather-cfg}
\end{equation}
where $s_{\mathrm{scene}}$ is kept low to avoid over-redrawing the source scene, $s_{\mathrm{weather}}$ controls the weather residual strength, and $\lambda_s$ is the same state scale used by UniSA. Thus Real samples suppress the weather residual, while Weather samples concentrate concrete weather changes in $\Delta_{\mathrm{weather}}$.

%% file: sections/5_experiments.tex
\section{Experiments}
\label{sec:experiments}

\subsection{Experimental Setup}
\label{subsec:eval-tracks}

\paragraph{Benchmark.}
We evaluate on a held-out benchmark split of HoloStateData, which contains 150 curated samples organized into Real, Simulation, and V2V subsets with 50 samples each. Following prior controllable video generation protocols that use curated evaluation sets to cover key motion and control factors~\citep{symphomotion,motionctrl}, our benchmark is stratified by source type and evaluation goal rather than sampled as a single mixed test pool. The Real subset contains unpaired samples whose origin and target are the same video, and evaluates whether the model follows camera and object controls for world preservation. The Weather subset combines the Simulation and V2V subsets for weather transfer, and evaluates whether the model can synthesize a target weather state from a single image, source-side camera/object controls, and a weather description. We report Real and Weather results separately because they test different sides of the source-to-state contract.

\vspace{-0.1in}
\paragraph{Baselines and input contract.}
For Real evaluation, we compare with Uni3C, GEN3C, VerseCrafter, and NeoVerse, which cover camera-aware and camera-object controllable video generation models. For Weather evaluation, we compare with Cosmos-Transfer2.5 and Wan2.7-Edit. These weather baselines are video-to-video editing models and can use a complete source video, whereas Holo-World uses only the first frame together with structured camera, object, and weather controls, we keep this input difference explicit.

\vspace{-0.1in}
\paragraph{Implementation details.}
Holo-World is built on Wan2.1-T2V-14B, we freeze the VAE, text encoder, and Wan backbone, and train only the World and State Adapter parameters in UniSA, following in~\citep{vace,wan}. UniSA is inserted at DiT layers $\{0,5,10,15,20,25\}$. Training uses 81-frame $480\times832$ video clips and a text length of 512. The global batch size is 16, learning rate is $2\times10^{-5}$, weight decay is 0.03, and we train models for 3,000 steps. At inference time, all methods generate 81-frame videos. We use 50 denoising steps, disables vanilla  CFG, sets $s_{\mathrm{scene}}=1.0$ and $s_{\mathrm{weather}}=2.0$ for SW-CFG.

\subsection{Comparison Results}
\label{subsec:comparison-results}

\paragraph{Results of World Preservation.}
We first evaluate the Real subset, where the model should follow camera and object controls while preserving the observed world. As shown in Table~\ref{tab-real-vbench}, Holo-World achieves the best overall VBench-I2V score compared to other methods, with strong I2V subject and background consistency. Some specialized baselines remain competitive on individual perceptual submetrics, which confirms that video quality alone is insufficient for evaluating unified state control. Table~\ref{tab-real-control} further shows that Holo-World obtains the lowest camera and object-control errors. Together, these results show that training for weather-state transfer does not prevent the model from preserving the source world under structured controls.

\begin{figure}[!t]
  \centering
  \includegraphics[width=\linewidth,height=0.25\textheight,keepaspectratio]{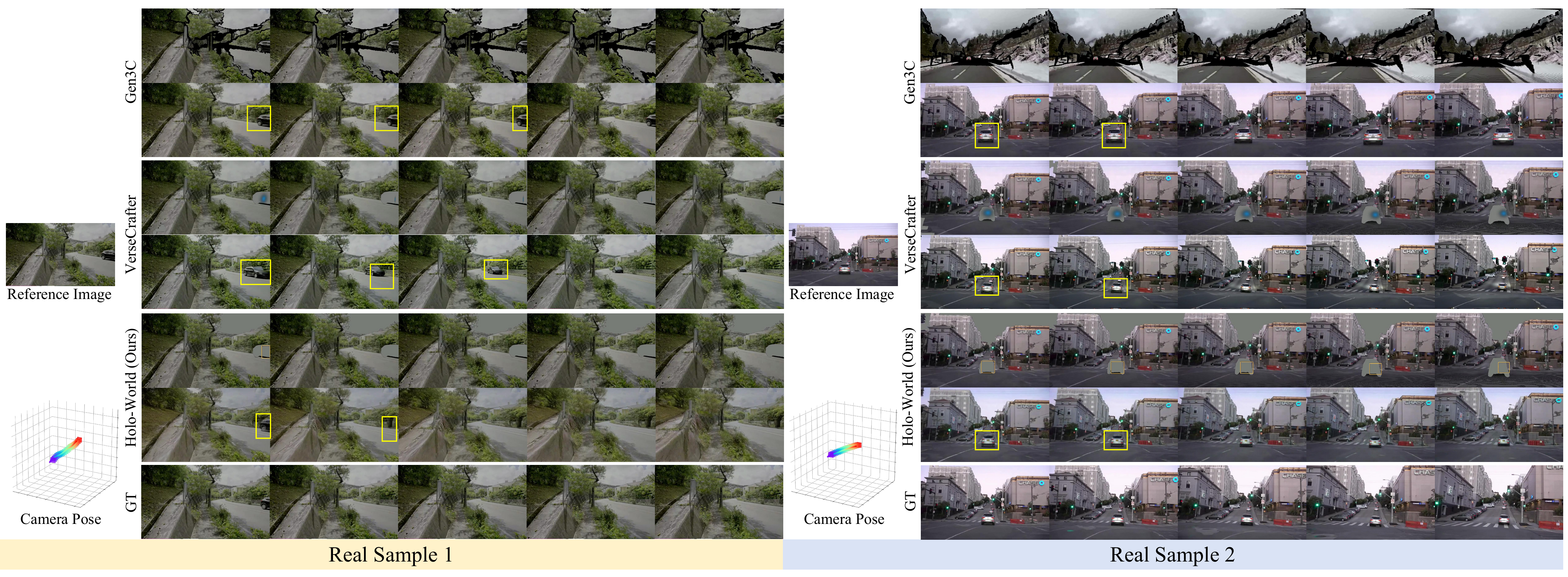}\\[-0.35em]
  \textbf{\small (a) World preservation.}\\[-0.05em]
  \includegraphics[width=\linewidth,height=0.25\textheight,keepaspectratio]{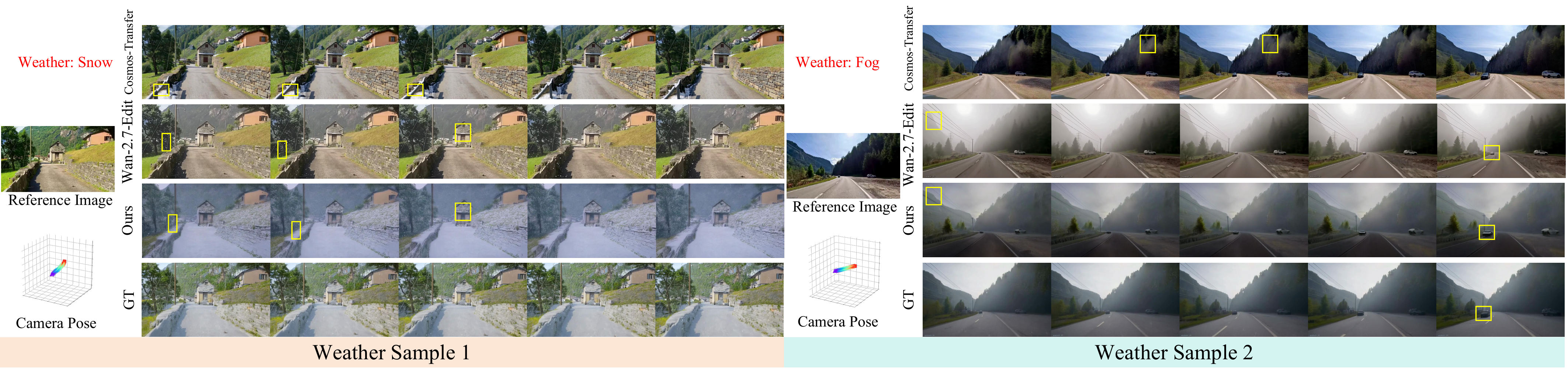}\\[-0.35em]
  \textbf{\small (b) Weather Transfer.}
  \caption{\small \textbf{Main qualitative comparison.} Real and Weather rows show the source-to-state requirement of preserving the source-controlled world while synthesizing the requested weather state.}
  \label{fig-main-comparison}
\end{figure}

\vspace{-0.1in}
\paragraph{Results of Weather Transfer.}
We next evaluate the Weather subset, where Holo-World must change the weather from a single image under camera and object control. This setting is stricter than other video-to-video editing models because Holo-World cannot inherit future layout, motion, or temporal continuity from a complete source video. Table~\ref{tab-weather-results} shows that Holo-World (I2V) outperforms other methods (V2V) on all weather metrics, reaching 86.00\% Weather Alignment and 68.51 VLM Evaluation. Human raters also prefer Holo-World in 83.00\% of comparisons against Cosmos-Transfer2.5 and 62.00\% against Wan2.7-Edit. The qualitative comparison in Figure~\ref{fig-main-comparison} mirrors these trends. The Real rows preserve the source-controlled world, and the Weather rows show clear weather-state transfer while maintaining consistent camera and object evolution.

\begin{table}[!t]
  \caption{\small \textbf{Quantitative comparison of different approaches on the Real subset under VBench-I2V evaluation.} The best results are highlighted in \textbf{bold}, and the second-best results are \underline{underlined}.}
  \label{tab-real-vbench}
  \centering
  \scriptsize
  \renewcommand{\arraystretch}{1.08}
  \setlength{\tabcolsep}{2.8pt}
  \resizebox{\linewidth}{!}{%
  \begin{tabular}{@{}lccccccccc@{}}
    \toprule
    \textbf{Method} &
    \makecell[c]{\textbf{Overall $\uparrow$}\\\textbf{Score}} &
    \makecell[c]{\textbf{Subject $\uparrow$}\\\textbf{Consistency}} &
    \makecell[c]{\textbf{Background $\uparrow$}\\\textbf{Consistency}} &
    \makecell[c]{\textbf{Motion $\uparrow$}\\\textbf{Smoothness}} &
    \makecell[c]{\textbf{Dynamic $\uparrow$}\\\textbf{Degree}} &
    \makecell[c]{\textbf{Aesthetic $\uparrow$}\\\textbf{Quality}} &
    \makecell[c]{\textbf{Imaging $\uparrow$}\\\textbf{Quality}} &
    \makecell[c]{\textbf{I2V $\uparrow$}\\\textbf{Subject}} &
    \makecell[c]{\textbf{I2V $\uparrow$}\\\textbf{Background}} \\
    \midrule
    Uni3C~\citep{uni3c} & 88.25 & 94.74 & 93.66 & 98.95 & \underline{56.00} & \textbf{50.12} & 70.45 & 98.45 & 98.58 \\
    GEN3C~\citep{gen3c} & 87.24 & \textbf{96.79} & \textbf{94.28} & \textbf{99.46} & 18.00 & 49.84 & 68.35 & 99.25 & \underline{99.30} \\
    VerseCrafter~\citep{versecrafter} & \underline{88.41} & 94.70 & \underline{93.93} & 98.96 & 46.15 & 48.83 & \textbf{72.87} & \underline{99.28} & 99.24 \\
    NeoVerse~\citep{neoverse} & 88.00 & 93.49 & 93.01 & \underline{99.43} & \textbf{58.00} & 49.39 & 65.16 & 98.96 & 98.93 \\
    Holo-World (Ours) & \textbf{89.05} & \underline{96.27} & 93.83 & 99.13 & 52.00 & \underline{50.00} & \underline{72.43} & \textbf{99.43} & \textbf{99.40} \\
    \bottomrule
  \end{tabular}%
  }
\end{table}

\begin{table*}[!t]
    \centering

    \begin{minipage}[t]{0.45\linewidth}
        \centering
        \caption{\small \textbf{Camera and object-control results on the Real subset.} The best results are highlighted in \textbf{bold}, and the second-best are \underline{underlined}.}
        \label{tab-real-control}

        \scriptsize
        \setlength{\tabcolsep}{1.6pt}
        \renewcommand{\arraystretch}{1.00}
        \begin{tabular*}{\linewidth}{@{\extracolsep{\fill}}lccc@{}}
            \toprule
            Method & RotErr $\downarrow$ & TransErr $\downarrow$ & ObjMC $\downarrow$ \\
            \midrule
            Uni3C~\citep{uni3c} & 3.036 & 1.228 & 7.771 \\
            GEN3C~\citep{gen3c} & 3.063 & \underline{1.156} & 8.078 \\
            VerseCrafter~\citep{versecrafter} & \underline{0.834} & 1.197 & \underline{7.002} \\
            NeoVerse~\citep{neoverse} & 1.703 & 2.112 & 8.859 \\
            Holo-World (Ours) & \textbf{0.719} & \textbf{1.123} & \textbf{6.549} \\
            \bottomrule
        \end{tabular*}
    \end{minipage}%
    \hfill%
    \begin{minipage}[t]{0.52\linewidth}
        \centering
        \caption{\small \textbf{Weather generation and user study results on the Weather subset.}
        The best results are highlighted in \textbf{bold}. User study reports blind pairwise preference rate against Cosmos-Transfer2.5 and Wan2.7-Edit.}
        \label{tab-weather-results}

        \scriptsize
        \setlength{\tabcolsep}{1.6pt}
        \renewcommand{\arraystretch}{0.92}
        \begin{tabular*}{\linewidth}{@{\extracolsep{\fill}}lccc@{}}
          \toprule
          Method &
          \makecell[c]{Weather $\uparrow$\\Alignment} &
          \makecell[c]{VLM $\uparrow$\\Evaluation} &
          \makecell[c]{User Study $\uparrow$\\Cosmos / Wan2.7} \\
          \midrule
          Wan2.7-Edit & 79.00 & 61.92 & -- \\
          Cosmos~\citep{cosmos} & 30.00 & 47.64 & -- \\
          Holo-World (Ours) & \textbf{86.00} & \textbf{68.51} & \textbf{83.00 / 62.00} \\
          \bottomrule
        \end{tabular*}
    \end{minipage}

\end{table*}

\vspace{-0.1in}
\subsection{Ablation Studies}
\label{subsec:ablations}

\begin{figure}[!t]
  \centering
  \includegraphics[width=0.96\linewidth,height=0.21\textheight,keepaspectratio]{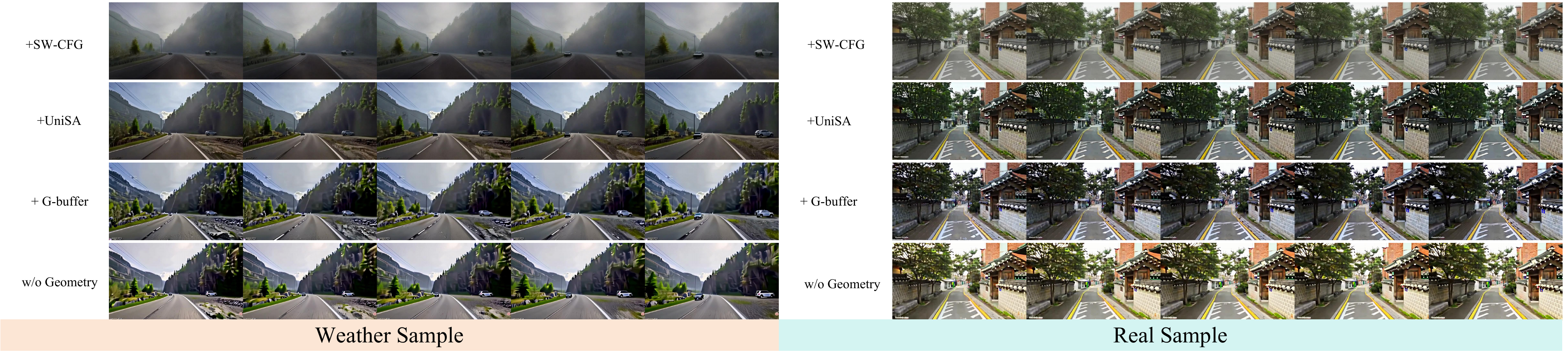}
  \caption{\textbf{Visualization of the model design ablation.} The comparison visualizes whether UniSA and SW-CFG reduce interference between Real-subset preservation and Weather-subset editing.}
  \vspace{-0.25em}
  \label{fig-adapter-ablation}

\end{figure}

\paragraph{Model design.}
We ablate the components that separate world preservation from weather transfer, including G-buffer controls, UniSA, and SW-CFG. World preservation is evaluated using background metrics on the Real subset, and weather transfer is evaluated using Weather Alignment and VLM Evaluation on the Weather subset. As shown in Table~\ref{tab-core-ablation}, adding G-buffer control improves Background PSNR and SSIM on the Real subset, and raise Weather Alignment and VLM Evaluation on the Weather subset, indicating that explicit geometry anchors help produce weather effects and preserve the controlled scene. UniSA further improves all three background metrics and VLM Evaluation, but the Weather Alignment score decreases slightly, suggesting that architectural separation alone can make the model conservative about visible cloud, rain, snow, or fog. With SW-CFG, the full model achieves the best background preservation and the strongest weather metrics at the same time. This pattern supports the main design claim that unified state control requires both separated residual subspaces and separated sampling guidance. Figure~\ref{fig-adapter-ablation} qualitatively confirms the same trend by comparing preservation and weather-transfer results under matched source controls.

\begin{figure}[!t]
  \centering
  \includegraphics[width=0.96\linewidth,height=0.21\textheight,keepaspectratio]{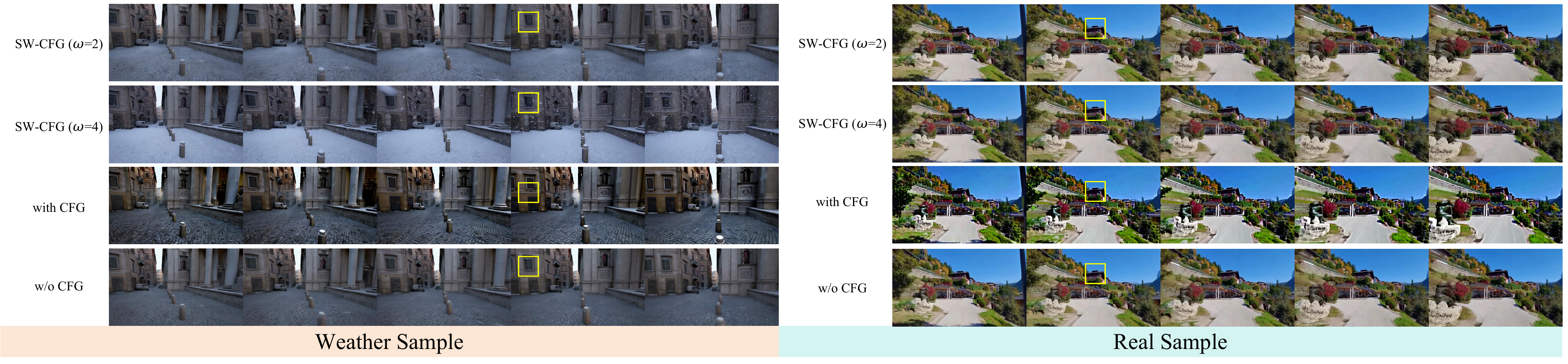}
  \caption{\textbf{Visualization of guidance ablation.} The comparison visualizes how no CFG, vanilla CFG, and Scene-Weather Decomposed CFG balance weather strength and source-world preservation.}
  \label{fig-cfg-ablation}
\end{figure}

\vspace{-0.1in}
\paragraph{Scene-Weather guidance.}
We further validate the sampling guidance ablations using the same trained UniSA model. Table~\ref{tab-cfg-ablation} shows that vanilla CFG improves weather metrics only modestly and degrades background preservation, which is consistent with the guidance conflict discussed in Section~\ref{subsec:cfg}. In contrast, Scene-Weather Decomposed CFG increases Weather Alignment and VLM Evaluation on the Weather subset while keeping the background metrics close to the no-CFG setting on the Real subset. This indicates that the decomposed weather residual strengthens cloud, rain, snow, and fog generation under the same camera, object, and geometry controls, instead of globally amplifying the full condition. Meanwhile, excessive weather guidance disrupts the scene structure (see $\omega=4$ in Figure~\ref{fig-cfg-ablation}), so we balance the weather scale settings in SW-CFG. Figure~\ref{fig-cfg-ablation} visualizes this trade-off, and Appendix~\ref{app:additional-qualitative} provides a detailed visual analysis of how SW-CFG balances weather-strength enhancement and source-world preservation. Appendix Figure~\ref{fig-multi-weather} provides multi-weather examples under fixed source-world. 
\vspace{-0.1in}

\begin{table*}[!t]
    \centering

    \begin{minipage}[t]{0.53\linewidth}
        \centering
        \caption{\textbf{Ablation on the model design.}
        We progressively enable G-buffer controls, UniSA, and SW-CFG to assess the individual contribution of different components to background preservation and weather controllability.}
        \label{tab-core-ablation}

        \scriptsize
        \setlength{\tabcolsep}{1.4pt}
        \renewcommand{\arraystretch}{1.05}
        \begin{tabular*}{\linewidth}{@{\extracolsep{\fill}}cccccccc@{}}
            \toprule
            G-Buffer & UniSA & SW-CFG &
            \makecell[c]{Back.\\PSNR $\uparrow$} &
            \makecell[c]{Back.\\SSIM $\uparrow$} &
            \makecell[c]{Back.\\LPIPS $\downarrow$} &
            \makecell[c]{Weather\\Align. $\uparrow$} &
            \makecell[c]{VLM\\Eval. $\uparrow$} \\
            \midrule
             &  &  & 13.20 & 0.552 & 0.320 & 51.00 & 51.60 \\
            \cmark &  &  & 15.34 & 0.577 & 0.329 & 62.00 & 55.20 \\
            \cmark & \cmark &  & 16.36 & 0.598 & 0.308 & 58.00 & 58.53 \\
            \cmark & \cmark & \cmark &
            \textbf{18.12} & \textbf{0.624} & \textbf{0.245} &
            \textbf{86.00} & \textbf{68.51} \\
            \bottomrule
        \end{tabular*}
    \end{minipage}%
    \hfill%
    \begin{minipage}[t]{0.44\linewidth}
        \centering
        \caption{\textbf{Ablation on Scene-Weather Decomposed CFG.}
        We compare no guidance, vanilla CFG, and different SW-CFG weather scales using the same trained UniSA model.}
        \label{tab-cfg-ablation}

        \scriptsize
        \setlength{\tabcolsep}{1.4pt}
        \renewcommand{\arraystretch}{1.05}
        \begin{tabular*}{\linewidth}{@{\extracolsep{\fill}}lccccc@{}}
            \toprule
            Method &
            \makecell[c]{Back.\\PSNR $\uparrow$} &
            \makecell[c]{Back.\\SSIM $\uparrow$} &
            \makecell[c]{Back.\\LPIPS $\downarrow$} &
            \makecell[c]{Weather\\Align. $\uparrow$} &
            \makecell[c]{VLM\\Eval. $\uparrow$} \\
            \midrule
            w/o CFG & 18.12 & 0.624 & 0.245 & 51.00 & 55.68 \\
            Vanilla CFG & 16.36 & 0.598 & 0.308 & 58.00 & 58.53 \\
            SW-CFG ($\omega$=2) & 18.12 & 0.624 & 0.245 & 86.00 & 68.51 \\
            SW-CFG ($\omega$=4) &
            \textbf{18.12} & \textbf{0.624} & \textbf{0.245} &
            \textbf{96.00} & \textbf{78.42} \\
            \bottomrule
        \end{tabular*}
    \end{minipage}

\end{table*}

\FloatBarrier

%% file: sections/6_conclusion.tex
\section{Conclusion}
\label{sec:conclusion}

We presented Holo-World, a controllable video world model for unified camera, object, and weather control from a single observed image. HoloStateData provides source-to-state supervision that separates source-world evidence from target-weather generation, while UniSA and Scene-Weather Decomposed CFG carry this separation into training and sampling. Experiments show that Holo-World preserves background and scene structure while transferring diverse weather states, outperforming V2V baselines on the Weather subset in an I2V setting.

%% file: sections/A_appendix.tex
\section{Appendix}
\raggedbottom

\subsection{Flow Matching Objective}
\label{app:prelim-flow-matching}

Flow Matching learns a time-dependent velocity field that transports samples between the data distribution and a simple prior distribution~\citep{flow_matching,ddpm,ddim}. Given $z_0$ as a clean video latent and $z_1\sim\mathcal{N}(0,I)$ as a Gaussian latent. For $\tau\in[0,1]$, the linear probability path is
\begin{equation}
z_{\tau}=(1-\tau)z_0+\tau z_1 .
\end{equation}
The corresponding velocity target is constant along this path:
\begin{equation}
u_{\tau}(z_0,z_1)=\frac{d z_{\tau}}{d\tau}=z_1-z_0 .
\end{equation}
The model therefore predicts the path velocity rather than directly predicting the clean latent. Holo-World uses Wan2.1-T2V-14B as the pretrained latent video backbone~\citep{wan}. Wan2.1 compresses videos into a latent space via a 3D VAE, encodes text conditions with a text encoder~\citep{umt5}, and uses a DiT denoiser~\citep{dit} to predict the flow target over spatio-temporal latent tokens. Holo-World keeps this Wan flow objective unchanged. Given conditioning $\mathcal{C}$, training minimizes
\begin{equation}
\mathcal{L}_{\mathrm{flow}}
=
\mathbb{E}_{z_0,z_1,\tau}
\left[
\left\|
v_{\theta}(z_{\tau},\tau,\mathcal{C})-(z_1-z_0)
\right\|_2^2
\right],
\end{equation}
where $\theta$ represents the frozen Wan VAE, text encoder, and DiT backbone. In Holo-World, $\mathcal{C}$ contains the first frame, source-side world controls, source appearance anchors, and factorized text conditions. UniSA only changes how these conditions enter selected Wan DiT blocks through trainable residual adapters, while the latent flow path and velocity objective remain those of the Wan2.1-T2V backbone.

\subsection{Additional Capability Comparison}
\label{app:capability-comparison}

Table~\ref{tab-capability-comparison} shows the capability comparison along the control axes exposed by each SOTA method. The goal is not to rank methods by generation quality, but to clarify their evidence contract. Existing image-to-video (I2V) methods expose camera control or joint camera-object control via 3D rendered geometry, while weather-oriented baselines are typically formulated as video-to-video (V2V) editing models that need input a complete source video. Holo-World targets the harder I2V task in which the model takes as inputs only a single image, explicit camera and object controls with geometry, and a target weather state. The model must synthesize future layout, motion, and weather-dependent appearance together, rather than relying on an already available future video as the editing substrate. This makes the weather subset results in Table~\ref{tab-weather-results} and the qualitative comparison in Figure~\ref{fig-main-comparison} especially demanding, since Holo-World outperforms the evaluated V2V weather-editing baselines while operating with weaker source evidence and a broader control interface.

\begin{table}[H]
  \caption{\textbf{Capability comparison of controllable video generation paradigms.} \cmark{} denotes an explicit control axis in the method's primary inference interface, and blank cells indicate that the corresponding state factor is not independently exposed.}
  \label{tab-capability-comparison}
  \centering
  \scriptsize
  \renewcommand{\arraystretch}{0.78}
  \setlength{\tabcolsep}{4.2pt}
  \begin{tabular}{lcccc}
    \toprule
    Method & Paradigm & Camera Control & Object Control & Weather Control \\
    \midrule
    GEN3C~\citep{gen3c} & I2V & \cmark & & \\
    VerseCrafter~\citep{versecrafter} & I2V & \cmark & \cmark & \\
    Uni3C~\citep{uni3c} & I2V & \cmark & \cmark & \\
    NeoVerse~\citep{neoverse} & I2V & \cmark & & \\
    Wan2.7-Edit & V2V & & & \cmark \\
    Cosmos-Transfer2.5~\citep{cosmos} & V2V & & & \cmark \\
    \textbf{Holo-World (Ours)} & I2V & \cmark & \cmark & \cmark \\
    \bottomrule
  \end{tabular}
  \vspace{-0.35em}
\end{table}
\vspace{-0.45em}
\FloatBarrier

\subsection{Additional Results}
\label{app:additional-qualitative}

This section provides additional results for the world preservation and weather transfer capabilities of our Holo-World. In all samples, the first frame, camera trajectory, rendered geometry controls, and object controls are defined by the source world, while the target-weather prompt specifies the environmental state to synthesize.

Figure~\ref{fig-cfg-ablation} further illustrates the guidance conflict that motivates SW-CFG. On the Weather sample, removing CFG preserves the source structure but produces weak weather changes, while vanilla CFG makes the target weather more visible at the cost of over-amplified appearance shifts and less stable scene structure. On the Real sample, stronger global guidance also changes the source appearance even though no weather transfer is requested, showing that a single CFG direction couples preservation and transfer. SW-CFG separates these two effects: the scene residual keeps the camera-conditioned world and object region stable, while the weather residual can be scaled independently to strengthen rain, snow, fog, or cloud effects. Increasing the weather scale therefore improves weather visibility on Weather samples without globally amplifying the full scene condition, which explains why SW-CFG better balances source-world preservation and weather-state transfer.

\begin{figure}[H]
  \centering
  \includegraphics[width=0.98\linewidth,height=0.30\textheight,keepaspectratio]{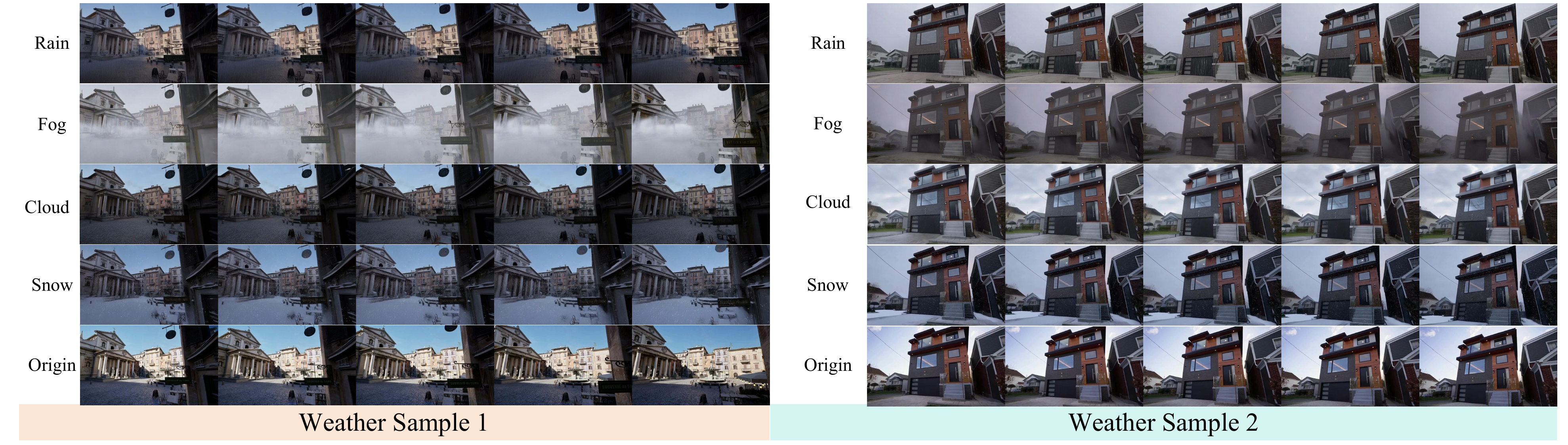}\\[0.35em]
  \includegraphics[width=0.98\linewidth,height=0.30\textheight,keepaspectratio]{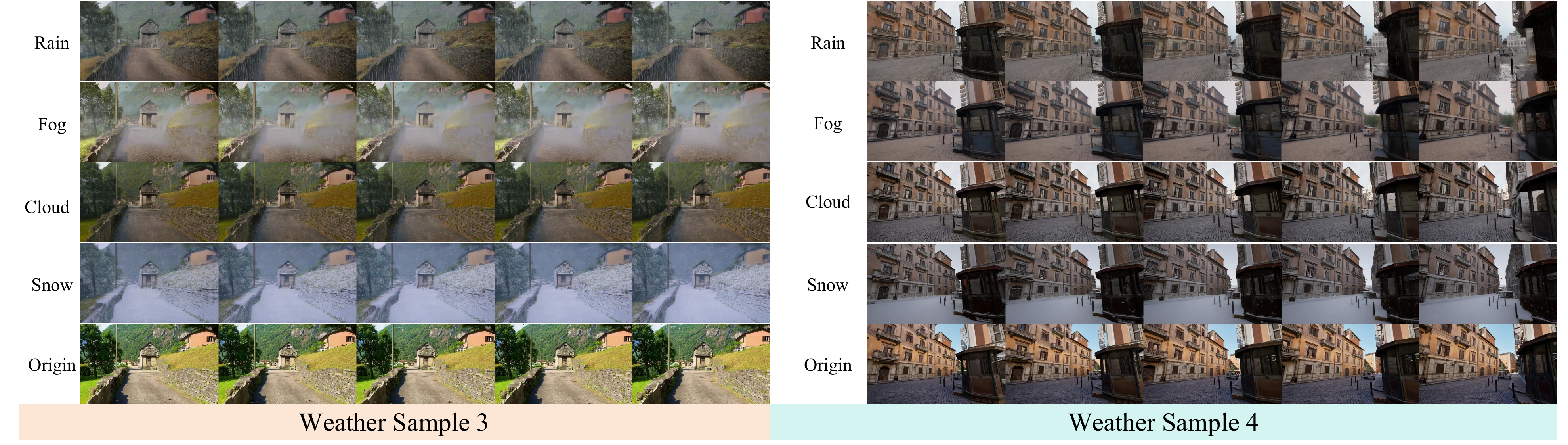}
  \caption{\textbf{Multi-weather state control.} With the same source world and camera/object controls, changing only the target weather prompt produces different weather videos.}
  \label{fig-multi-weather}
\end{figure}

Figure~\ref{fig-multi-weather} evaluates multi-weather results under fixed source-world controls. Across urban streets, residential scenes, mountain roads, and city plazas, changing the weather prompt produces clearly different rain, fog, cloud, and snow states while retaining the same viewpoint progression and scene layout. Rain variants darken the scene and introduce wet or rainy appearance, fog variants reduce long-range visibility and soften distant structures, cloud variants shift the illumination toward overcast conditions, and snow variants alter ground and surface appearance. The repeated temporal columns remain aligned within each sample, showing that weather transfer does not override the underlying camera-conditioned world evolution.

\begin{figure}[H]
  \centering
  \includegraphics[width=\linewidth,height=0.25\textheight,keepaspectratio]{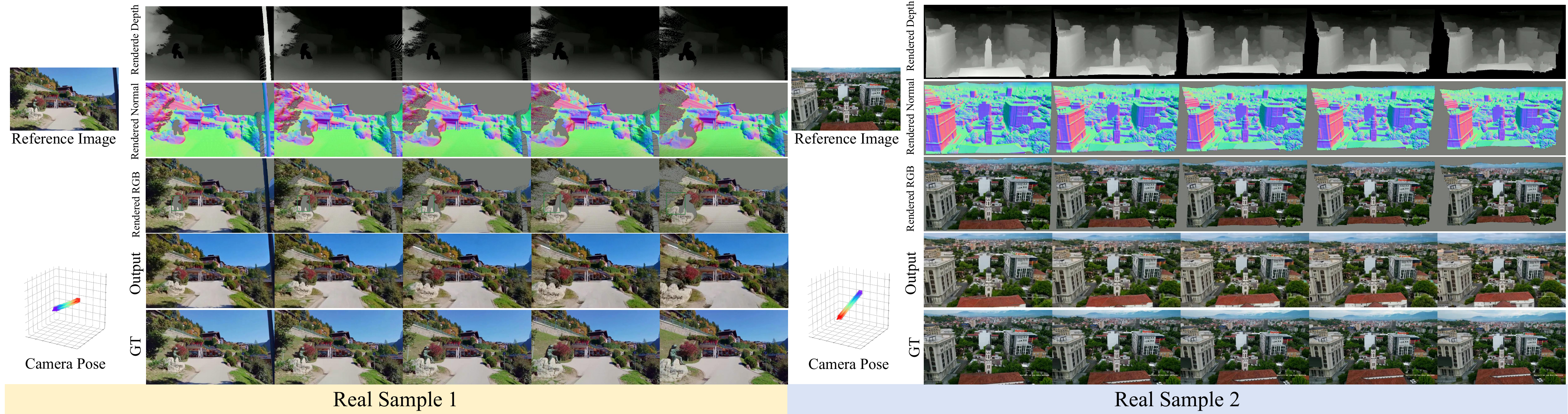}\\[0.18em]
  \textbf{(a) World preservation.}\\[0.45em]
  \includegraphics[width=\linewidth,height=0.25\textheight,keepaspectratio]{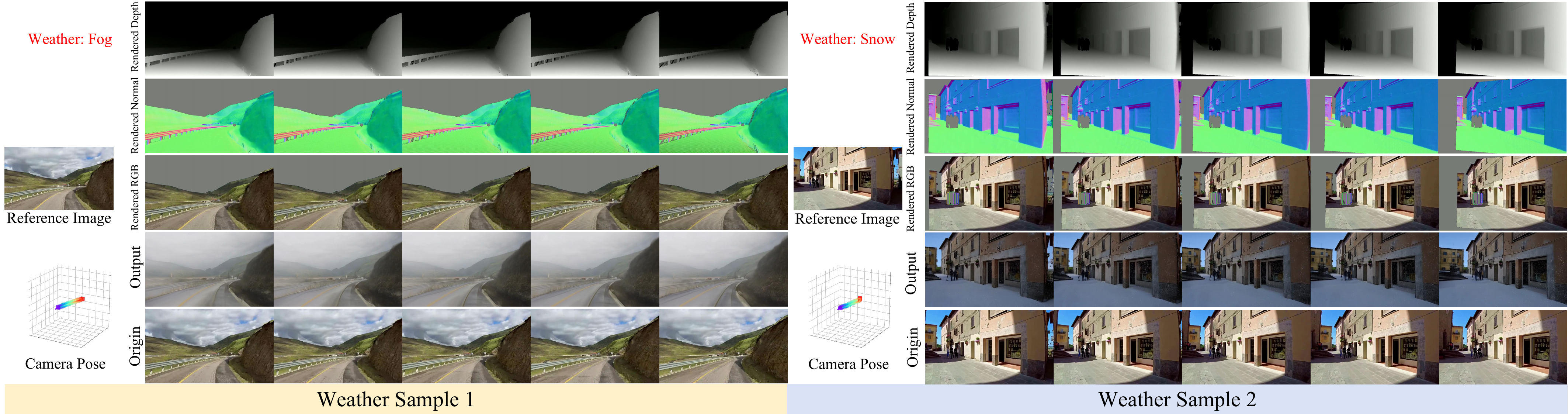}\\[0.18em]
  \textbf{(b) Weather transfer.}
  \caption{\textbf{Additional visualization of Holo-World.} Real examples visualize background-consistent generation under rendered controls, while Weather examples visualize target weather transfer under the same source-side control interface.}
  \label{fig-appendix-visualization}
\end{figure}

Figure~\ref{fig-appendix-visualization} exposes the intermediate representations used by our Holo-World and compares to the generated videos. In the Real examples, rendered RGB supplies the camera-conditioned background, while depth and normal buffers provide geometry-aligned conditions for scene structure. The outputs follow the rendered controls across time, preserving road boundaries, building facades, vegetation layout, and object placement without introducing unnecessary weather or global restyling. This demonstrates that our Holo-World can preserve the world as expected by the source-side controls.

The Weather examples use the same type of source-side controls but with a target weather state. In the fog case, Holo-World lowers distant contrast and visibility along the road while preserving the road curvature, guardrail structure, mountainside boundary, and camera progression. In the snow case, the model changes the street and facade appearance toward a snowy state while maintaining the original building geometry and viewpoint trajectory. The weather effects remain spatially grounded in the rendered scene structure rather than appearing as drifting texture overlays. Figures~\ref{fig-multi-weather} and~\ref{fig-appendix-visualization} show that Holo-World uses source controls to determine which world should be followed and uses the weather condition to determine how that world should be rendered.
Table~\ref{tab-weather-control-fidelity} further reports the same camera and object-control metrics on the Weather subset. This diagnostic complements the Real-subset control results in Table~\ref{tab-real-control} and verifies that Holo-World still preserves camera and object control fidelity while changing the scene into a target weather state.

\FloatBarrier

\begin{table}[H]
  \caption{\textbf{Camera and object-control results on the Weather subset.} The metrics follow the same RotErr, TransErr, and ObjMC definitions as Table~\ref{tab-real-control}, but are computed on Simulation and V2V weather-transfer samples.}
  \label{tab-weather-control-fidelity}
  \centering
  \scriptsize
  \renewcommand{\arraystretch}{0.86}
  \setlength{\tabcolsep}{7pt}
  \begin{tabular}{lccc}
    \toprule
    Method & RotErr $\downarrow$ & TransErr $\downarrow$ & ObjMC $\downarrow$ \\
    \midrule
    Holo-World & 1.681 & 1.745 & 6.744 \\
    \bottomrule
  \end{tabular}
\end{table}

\subsection{Application to nuScenes Driving Scenes}
\label{app:nuscenes}

\begin{figure}[H]
  \centering
  \includegraphics[width=\linewidth]{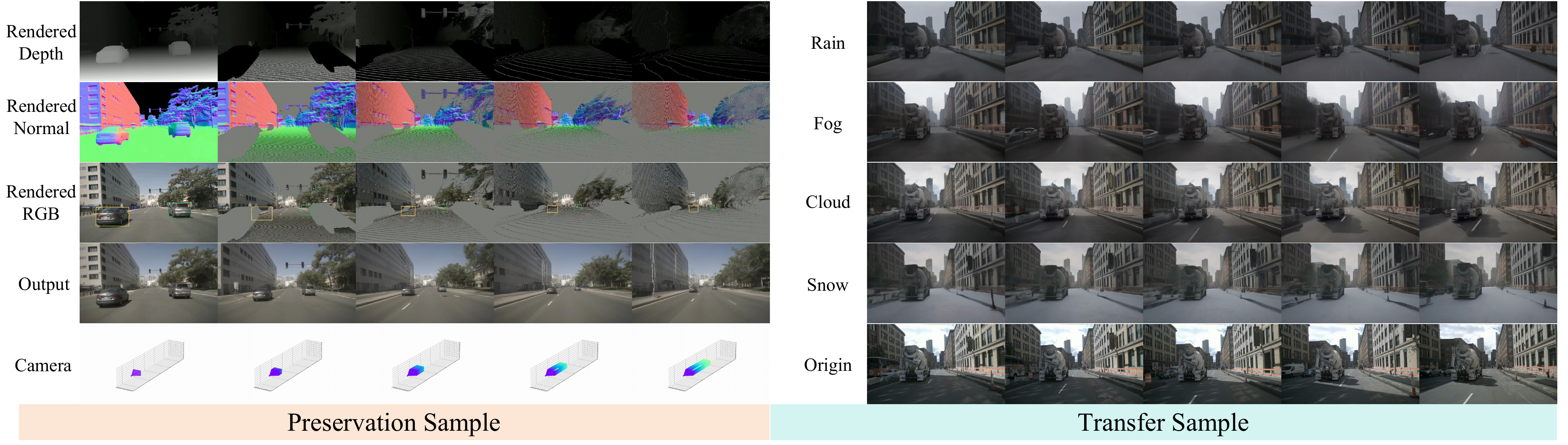}
  \caption{\textbf{Application of Holo-World to nuScenes driving scenes.} Holo-World supports world preservation (left) and weather transfer (right) under source-side camera and object controls. The original video is shown for reference in the transfer example. }
  \label{fig-appendix-nuscenes}
\end{figure}

To explore Holo-World in autonomous-driving scenes, we extend HoloStateData with additionally annotated video clips from the nuScenes dataset and include them in Holo-World training. These clips follow the same source-to-state annotation format, with camera trajectories, object controls, and rendered RGB, depth, and normal buffers. Figure~\ref{fig-appendix-nuscenes} shows qualitative results for world preservation and weather transfer on the nuScenes dataset. In the preservation example, the generated sequence follows the forward camera motion while retaining the overall road layout, surrounding buildings, and leading vehicles, despite increasingly incomplete source renderings. In the transfer example, the same source scene is generated under rain, fog, cloud, and snow conditions: rain darkens the scene and introduces visible rain streaks, fog reduces distant visibility, cloud changes the sky appearance, and snow covers the road surface. Across these variants, the street layout and leading truck remain broadly recognizable, although fine vehicle and facade details can still vary. These examples illustrate the applicability of Holo-World's unified control interface to driving scenes after incorporating nuScenes training data.

\subsection{HoloStateData Details}
\label{app:dataset-snapshot}

Figure~\ref{fig-holostatedata-example} illustrates how HoloStateData converts a raw video clip into a source-to-state training sample. Each sample starts from a source frame and source-side controls, including rendered background RGB, geometry buffers, camera motion, and object controls. These fields specify the world that should be followed during generation. When paired weather supervision is available, the sample additionally provides a target-weather video and a weather prompt condition, which specify how the same controlled world should be rendered under a new environmental state.

The key design is to keep source controls and target supervision separated. Rendered background controls and object controls are always constructed from the source world, so target-weather pixels cannot leak into the model-facing controls. Real samples use the original video as no-edit supervision for world preservation, while paired weather samples use the target-weather video only as the supervised output for weather. This shared schema lets HoloStateData train both background-consistent generation and weather-state transfer without changing the inference interface.

\begin{figure}[H]
  \centering
  \includegraphics[width=0.52\linewidth,height=0.28\textheight,keepaspectratio]{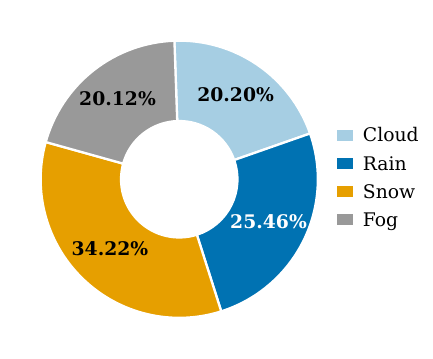}
  \caption{\textbf{Weather-family distribution in HoloStateData.} The pie chart shows the target weather-family distribution of the Weather training subset used for state-transfer supervision.}
  \label{fig-holostatedata-weather-distribution}
\end{figure}

\paragraph{Dataset split statistics.}
Table~\ref{tab-holostatedata-split} summarizes the split design of HoloStateData. The training set contains about 15K samples organized into Real, Simulation, and V2V subsets, while the held-out benchmark contains 150 samples with 50 samples from each subset so that no-edit world preservation and target-weather generation can be evaluated separately. The benchmark is designed as a stratified diagnostic split rather than a randomly sampled mixed test pool, so each subset corresponds to a distinct source-to-state evaluation contract. The Weather training subset is formed by the Simulation and V2V subsets and covers four target weather families. Figure~\ref{fig-holostatedata-weather-distribution} visualizes the family-level distribution, making the balance of cloud, rain, snow, and fog supervision explicit for training.

\begin{table}[H]
  \caption{\textbf{HoloStateData split statistics.} The benchmark is balanced across Real, Simulation, and V2V subsets, while training-set subset counts are reported for dataset transparency.}
  \label{tab-holostatedata-split}
  \centering
  \scriptsize
  \renewcommand{\arraystretch}{0.86}
  \setlength{\tabcolsep}{8pt}
  \begin{tabular}{lcc}
    \toprule
    Subset & Training Samples & Benchmark Samples \\
    \midrule
    Real & 7,571 & 50 \\
    Simulation & 3,541 & 50 \\
    V2V & 3,954 & 50 \\
    \midrule
    Total & 15,066 & 150 \\
    \bottomrule
  \end{tabular}
\end{table}

\paragraph{VLM text annotation.}
HoloStateData uses Qwen3-VL to produce factorized text conditions with low-randomness decoding~\citep{qwen3vl}. The annotation pipeline uses two VLM prompts rather than one generic caption. The scene prompt is generated from the source video and describes scene identity, spatial layout, visible objects, object relations, temporal evolution, and camera motion. It explicitly excludes weather, global illumination, imagined content, and hidden causes, so the source-world description does not bind the scene to a particular environmental state. For paired weather samples, a separate weather prompt is generated from the target-weather video and restricted to a compact weather taxonomy. Real samples keep the weather field empty. This factorized annotation keeps scene identity, camera/object evolution, and weather-state transfer in separate text channels.

\begin{tcolorbox}[enhanced,colback=blue!2,colframe=blue!45!black,boxrule=0.5pt,arc=1mm,left=1mm,right=1mm,top=1mm,bottom=1mm]
\begin{promptverbatim}
  You are a video caption generation specialist whose goal is to create high-quality, detailed English captions for the user's input video by closely observing its content. The captions are intended for video world model training data: prioritize scene layout, how scene state evolves over time, and how people and objects interact with the environment and each other, while still covering people and camera work when visible. Your task is to describe the video comprehensively and expressively while maintaining authenticity and naturalness. Integrate visual and temporal elements strictly adhering to the formatting of the examples provided.

  Task Requirements:
  1. Describe the main theme and setting (such as location and spatial layout of the scene). Do not describe weather or lighting here; omit those entirely (they are covered by a separate annotation prompt).
  2. Describe the environment and scene-anchored cues (terrain, architecture, furniture, obstacles, open space, notable static props), then key subjects (appearance, clothing, expression, quantity, ethnicity, posture, etc.), spatial relationships, and camera movements;
  3. Describe relationships and interactions between people, objects, and the environment (positioning, contact, support, occlusion, entry and exit, use of tools or furniture, etc.);
  4. Describe any character present and their emotions or activities (such as expressions, postures, etc.), in relation to nearby objects and the scene;
  5. The caption should reflect the style of the video (e.g., cinematic, anime, documentary, vlog, vintage film, etc.);
  6. Describe the content in chronological order, including scene-level changes (rearranged or moved objects, reconfiguration of space) and how characters or objects move; describe how the camera changes perspective. Use accurate verbs to describe movement;
  7. Describe background and environmental details (such as architecture, natural scenery, materials and clutter, etc.) without weather or lighting;
  8. Describe camera motion control (e.g., zoom in, zoom out, push in, pull out, pan right, pan left, truck right, truck left, tilt up, tilt down, pedestal up, pedestal down, arc shot,  tracking shot, static shot, and handheld shot, etc.);
  9. Do not describe imagined content. Only describe what can be determined from the video. Avoid listing things. Do not use abstract concepts (love, hate, justice, infinity, joy) as subjects. Use concrete nouns (human, cup, dog, planet, headphones) for more accurate results. Use verbs to describe movement and state changes. Write in plain, conversational language. Start with the scene or central situation when it is clear (often a place or situational phrase); otherwise start with the clearest main subject. Without "\n", subheading and title.
  10. Control the caption length to around 80-150 words;
\end{promptverbatim}
\end{tcolorbox}
\vspace{-0.35em}
{\footnotesize\noindent\textbf{(1) Scene annotation prompt.} The prompt extracts source-scene content and motion while excluding weather from the scene text.\par}
\vspace{0.5em}

\begin{tcolorbox}[enhanced,colback=blue!2,colframe=blue!45!black,boxrule=0.5pt,arc=1mm,left=1mm,right=1mm,top=1mm,bottom=1mm]
\begin{promptverbatim}
You are a weather-only caption specialist for video weather editing data. From the input video (or the specified target), produce one short English line that states only the weather condition using the fixed phrase inventory below. Do not describe scene layout, objects, people, lighting nuances, camera paths, or style--except where the chosen output template already mentions preserving scene or camera.

Weather taxonomy (pick exactly one family per output; never combine families):

1. Cloud (cloud-only) -- use one phrase only:
    - `a few clouds` / `cloudy` / `overcast`

2. Rain (precipitation and/or wet ground, independent) -- choose what is visibly present; do not force a pair. Use only phrases from the lists below:
    - Rain (precipitation): `light rain` / `rain` / `heavy rain`
    - Puddle / rain ground: `slightly wet rain ground` / `rain ground` / `flooded rain ground`
    - Compose one line: one matching phrase or, if both rain and wet ground are clearly present, two phrases in order `{rain phrase} and {puddle phrase}`. If only rain or only wet ground is visible, output that single phrase (no invented second tier).

3. Snow (falling snow and/or snow on ground, independent) -- same independence as rain: do not require falling snow and ground snow together. Use only:
    - Falling snow: `light snow` / `snow` / `heavy snow`
    - Snow-covered ground: `light snow-covered ground` / `snow ground` / `deep snow ground`
    - One phrase, or both as `{falling-snow phrase} and {ground-snow phrase}` when both are clearly visible; otherwise a single phrase matching what you see.

4. Fog (fog-only) -- use one phrase only:
    - `light fog` / `fog` / `thick fog`

Disambiguation rules:
- Do not add cloud or fog when the family is Rain or Snow, and do not add rain or snow when the family is Cloud or Fog. This keeps labels disjoint for training (cloudy / rainy / snowy / foggy are separate regimes).
- Do not mix rain and snow in one line; pick one dominant family. If both are strongly visible, prefer Rain for this dataset unless snow clearly dominates the frame.
- Use only the strings above for weather content--no synonyms (no "drizzle", "mist", "partly cloudy", etc.).

Output templates (choose exactly one per sample; do not output both):

- Template A: `Edit the video to show {phrases}. Keep the scene structure unchanged and only change the weather realistically.`

- Template B: `{phrases}. Realistic weather, preserve scene layout and camera motion.`

Replace `{phrases}` with the composed weather content for the chosen family: a single cloud or fog phrase; or for rain/snow, one or two inventory phrases joined by `and` only when both tiers are justified (rain before puddle; falling snow before ground snow).

Formatting:
- One single line of English; no line breaks, no subheadings, no quotes around the full line.
- Keep it short; only the template sentence(s) above.

Directly output the single English line and nothing else.
\end{promptverbatim}
\end{tcolorbox}
\vspace{-0.35em}
{\footnotesize\noindent\textbf{(2) Weather annotation prompt.} The prompt maps target-weather evidence to a compact weather-state phrase without adding scene, object, or camera descriptions.\par}

\paragraph{Automatic construction.}
The construction pipeline turns the same sample into model-facing controls. For the V2V subset, SpatialVID-HQ videos are used as source clips. Our weather-transfer model is fine-tuned from Wan2.1-Fun-Control-14B on open-source paired simulated weather videos~\citep{weatherweaver} rendered with Unreal Engine, and is used together with proprietary video editing models to generate candidate target-weather clips for SpatialVID-HQ sources. We retain generated pairs only when the target weather is clearly visible, the source scene layout is preserved, and no severe temporal artifacts or object-level distortions are observed. These checks are applied before the retained V2V samples enter the same source-to-state construction pipeline as the Real and Simulation subsets. Depth Anything and UniDepth provide monocular depth priors, MegaSaM estimates camera parameters, and MoGe supplies dense depth and normal anchors for rendering source-side background RGB and geometry-buffer controls along the camera trajectory~\citep{depthanythingv2,unidepth,megasam,droid_slam,moge}. For object control, Grounding DINO detects first-frame object candidates and SAM-2 propagates masks through the clip, after which each mask is converted into a tight per-frame bounding box~\citep{groundingdino,sam2}. The first rendered frame is pinned to the observed source frame and its geometry estimates to reduce conditioning-frame drift. For paired weather samples, the target-weather video is used only as the supervised output and never as a source for rendered controls. This construction rule matches the model design: world controls define the scene scaffold to preserve, while weather text and paired targets supervise the state in which that scaffold should be rendered.

\begingroup
\begin{figure}[H]
  \centering
  \includegraphics[width=\linewidth,height=0.42\textheight,keepaspectratio]{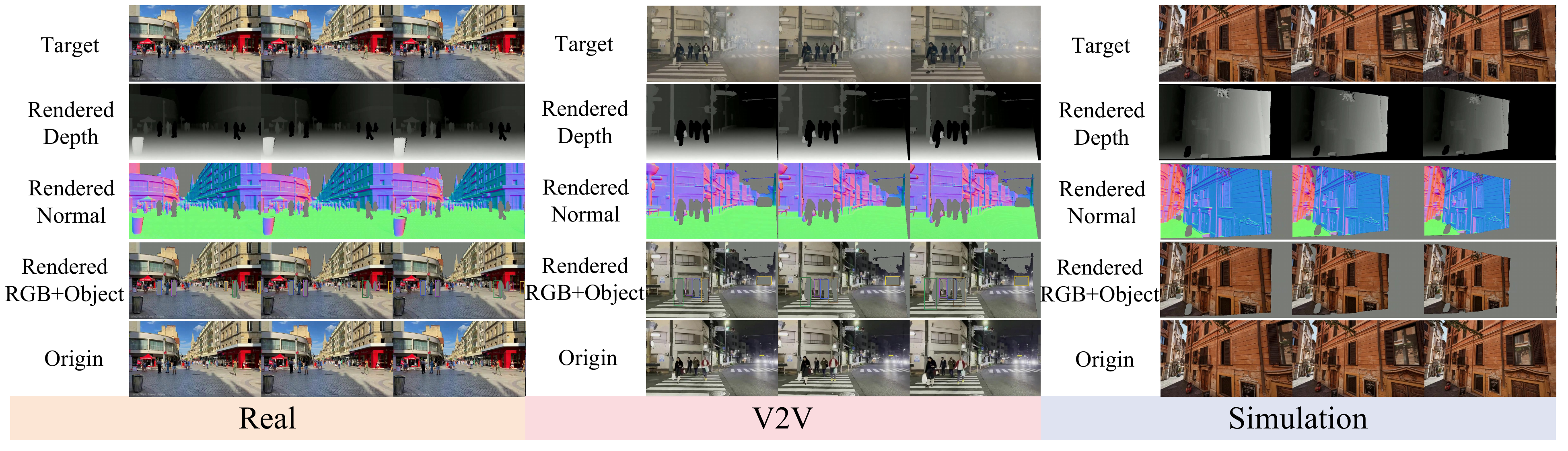}
  \caption{HoloStateData example. The visualization shows how one video record is converted into the source first frame, rendered world controls, object controls, and weather-state annotations used by the unified source-to-state training interface.}
  \label{fig-holostatedata-example}
\end{figure}
\endgroup
\FloatBarrier

\subsection{Evaluation of Holo-World}
\label{app:metrics}

\subsubsection{Metric Overview}
\label{app:metric-overview}

This appendix gives the formulas, prompts, evaluator caveats, and user-study protocol for VBench-I2V, camera/object-control errors, Weather Alignment, VLM Evaluation, and User Study. A valid Weather-subset run fixes the split, generated-video manifest, evaluator prompts, frame sampling, and human-study design before scoring any method. The Real subset uses standard video, camera, object, and rendered-background metrics; the Weather subset uses weather-specific metrics for target-weather semantics, generation quality, and human preference.

Weather Alignment ignores scene identity, object identity, camera motion, and general image quality except when they make the weather evidence unreadable. VLM Evaluation is broader: it asks whether the weather background, weather dynamics, and video quality are plausible, but it still does not score camera-control or object-control accuracy. User Study is the complementary human signal and is reported as two separate pairwise win rates rather than a single mixed score. For the weather metrics via VLM, we adopted GPT model to evaluate the Weather Alignment and VLM Evaluation.

Each Weather-subset item contains $I_0$, a target weather prompt $c_{\mathrm{state}}$, generated video $\hat{V}$, and optional source/control context for human raters. Automatic Weather Alignment and VLM Evaluation use only $c_{\mathrm{state}}$ and $\hat{V}$; the input context is reserved for provenance and human preference. Only Simulation and V2V subset samples participate because Real samples have no target weather edit. The three Weather-subset metrics are therefore never averaged with Real-subset scores or collapsed into a mixed benchmark number.

\subsubsection{VBench-I2V Metrics}
\label{app:vbench-i2v}

We evaluate image-to-video generation quality on the Real subset using the official VBench-I2V metrics~\citep{vbench++}. For each generated clip, the evaluator receives the conditioning image and the generated video, then reports eight dimensions used in the main table: imaging quality, aesthetic quality, dynamic degree, motion smoothness, background consistency, subject consistency, I2V subject consistency, and I2V background consistency, with camera motion excluded. Following VBench-I2V, the Overall Score is a weighted aggregation of these eight normalized dimension scores. Let $\{s_k\}_{k=1}^{8}$ denote the eight raw VBench-I2V scores
over the Real subset. We compute
\begin{equation}
S_{\mathrm{Overall}}
=
\frac{\sum_{k=1}^{8} w_k\,\operatorname{Norm}_k(s_k)}
     {\sum_{k=1}^{8} w_k}.
\label{eq:vbench-i2v}
\end{equation}
Here, $w_k$ denotes the effective per-dimension weight that preserves the within-group weighting and the equal weighting of the video-quality and I2V consistency groups after excluding camera motion. The main table reports the averaged VBench-I2V scores over the Real subset, with all values shown on a percentage scale.

\subsubsection{Camera and Object-Control Metrics}
\label{app:control-metrics}

We evaluate camera and object-control fidelity on the Real subset by comparing trajectories recovered from generated videos with the target trajectories. For each generated video, the evaluation methods estimate a camera trajectory and dynamic object trajectories using the same geometry annotation.

For camera control, let $\{R_{\mathrm{target}}^t,T_{\mathrm{target}}^t\}_{t=1}^{T}$ represent the target camera rotation and translation parameters and $\{R_{\mathrm{gen}}^t,T_{\mathrm{gen}}^t\}_{t=1}^{T}$ represent the camera trajectory estimated from the generated video. We calculate rotation error (RotErr) by the average geodesic distance on $\mathrm{SO}(3)$:
\begin{equation}
\mathrm{RotErr}
=
\frac{1}{T}
\sum_{t=1}^{T}
\arccos
\left(
\frac{
\mathrm{tr}(R_{\mathrm{gen}}^t (R_{\mathrm{target}}^t)^{\top})-1
}{2}
\right).
\label{eq:roterr}
\end{equation}
Translation error (TransErr) is the average Euclidean distance between the target and generated camera translations:
\begin{equation}
\mathrm{TransErr}
=
\frac{1}{T}
\sum_{t=1}^{T}
\left\|
T_{\mathrm{gen}}^t-T_{\mathrm{target}}^t
\right\|_2 .
\label{eq:transerr}
\end{equation}

For object motion control, let $N$ be the number of target controlled objects. For target object $o$ and predicted object $k$, their trajectory distance is
\begin{equation}
d(o,k)
=
\frac{1}{T}
\sum_{t=1}^{T}
\left\|
\hat{\mu}_{k}^{t}-\mu_{o}^{t}
\right\|_2 ,
\label{eq:obj-distance}
\end{equation}
where $\mu_{o}^{t}$ and $\hat{\mu}_{k}^{t}$ are the target and generated 3D object centers at frame $t$. Since object identities are not known in generated videos, we match target and predicted trajectories with the Hungarian algorithm. Unmatched target objects are assigned a fixed penalty $\lambda$. The object motion control score is
\begin{equation}
\mathrm{ObjMC}
=
\frac{1}{N}
\sum_{o=1}^{N}
d_o,
\qquad
d_o
=
\begin{cases}
d(o,k), & \text{if target object } o \text{ is matched to prediction } k,\\
\lambda, & \text{if } o \text{ is unmatched}.
\end{cases}
\label{eq:objmc}
\end{equation}
It is noted that we only calculate the special object class set "person . human . rider . bicycle . motorcycle . scooter . car . truck . bus . animal" via Grounded-SAM2, which is the same as the dataset annotation setting.

\subsubsection{Weather Alignment}
\label{app:weather-alignment}

Weather Alignment is a binary weather-semantics score. Its input is a generated video $\hat{V}$, the target weather prompt $c_{\mathrm{state}}$, and a frozen frame/clip sampler. Its output is one aligned/not-aligned decision per sample. For sample $i$,
\begin{equation}
a_i = E_{\mathrm{wa}}(\hat{V}^{(i)},c_{\mathrm{state}}^{(i)}),\qquad a_i\in\{0,1\},
\end{equation}
where $a_i=1$ denotes alignment. We report the pass rate
\begin{equation}
S_{\mathrm{WA}}
=
\frac{100}{N}
\sum_{i=1}^{N}
a_i.
\label{eq:weather-alignment}
\end{equation}
The evaluator prompt is intentionally short and exclusionary: it asks whether the target core weather family is clearly visible and forbids using scene identity, object identity, camera motion, composition, or general image quality as scoring criteria.

\begin{tcolorbox}[enhanced,colback=blue!2,colframe=blue!45!black,boxrule=0.5pt,arc=1mm,left=1mm,right=1mm,top=1mm,bottom=1mm]
\begin{promptverbatim}
You are evaluating only whether a generated video aligns with the target weather prompt.

Target weather prompt:
{c_state}

Only consider the core weather meaning in the prompt. The allowed core weather
types are: cloud, rain, snow, fog.

Generated video:
{video}

Question:
Does the generated video align with the target weather prompt?

Judge only whether the core weather requested by the prompt is visible:
- cloud: cloudy or overcast sky;
- rain: visible falling rain;
- snow: visible falling snow;
- fog: visible fog, haze, or reduced visibility caused by fog.

Do not judge scene identity, object identity, camera motion, composition,
or general video quality unless they make the weather condition impossible to judge.
Do not consider puddles, wet ground, snow accumulation, snow-covered ground,
weather intensity words, or any other secondary weather effects.

Return a JSON object:
{
  "aligned": true or false,
  "reason": "one short sentence explaining the weather evidence"
}

Decision rule:
aligned = true only if the core weather requested by the prompt is clearly visible.
aligned = false if the target weather is absent, ambiguous, too weak,
or belongs to another category.
\end{promptverbatim}
\end{tcolorbox}
\vspace{-0.35em}
{\footnotesize\noindent\textbf{(3) Weather alignment evaluator prompt.} The prompt measures only whether the target core weather family is visible and excludes scene, object, camera, and general quality factors.\par}
\vspace{0.5em}

The Weather-subset table reports the aligned-sample percentage over the 100 Simulation/V2V samples. Intensity words such as light rain, heavy snow, or thick fog do not change the criterion: WA only requires the core weather family to be visible.

\paragraph{Evaluator validation and bias.}
If Qwen3-VL is used for both annotation and evaluation, WA may inherit vocabulary and calibration bias. We therefore validate it on GPT model, distinguishing labeled VLMs.

\subsubsection{VLM Evaluation}
\label{app:vlm-evaluation}

VLM Evaluation scores weather generation quality on a 0--100 scale. The evaluator receives the target weather prompt and generated video, but not the target-weather video. It focuses on three subdimensions: weather background editing, weather dynamics, and video quality. This score is distinct from Weather Alignment: a video can pass Weather Alignment by showing visible rain or snow, yet receive a low VLM Evaluation score if the weather background is unchanged, particle motion is unnatural, or the video flickers. For sample $i$,
\begin{equation}
q_i = E_{\mathrm{vlm}}(\hat{V}^{(i)},c_{\mathrm{state}}^{(i)}),\qquad q_i\in[0,100],
\end{equation}
and the reported score is
\begin{equation}
S_{\mathrm{VLM}}
=
\frac{1}{N}\sum_{i=1}^{N} q_i.
\label{eq:vlm-evaluation}
\end{equation}

\begin{tcolorbox}[enhanced,colback=blue!2,colframe=blue!45!black,boxrule=0.5pt,arc=1mm,left=1mm,right=1mm,top=1mm,bottom=1mm]
\begin{promptverbatim}
  You are evaluating the quality of a generated weather video for a research benchmark.

  Target weather prompt:
  {c_state}
  
  Generated video:
  {video}
  
  Please evaluate only the generated weather result. Give one overall score from 0 to 100.
  Focus on:
  1. Weather Background: whether the sky, ground, road, buildings, visibility,
     surface materials, reflections, snow coverage, wetness, fog atmosphere,
     or other background cues are plausibly edited for the target weather.
  2. Weather Dynamics: whether dynamic weather elements such as rain, snow,
     fog, mist, or moving atmospheric effects appear natural, temporally coherent,
     and consistent with the target weather.
  
  Do not reward a method merely for keeping the background unchanged.
  A good result should edit the weather background when the target weather requires it.
  Do not evaluate camera-control accuracy or object-control accuracy.
  
  Use the following rough scoring scale:
  - 0-20: unusable result, no meaningful target weather generation, or weather is almost entirely absent.
  - 21-40: very weak weather generation; only minor or partial cues such as slight wetness, faint haze, or limited surface change.
  - 41-60: recognizable target weather, but background editing or weather dynamics are incomplete, weak, unstable, or unnatural.
  - 61-80: good weather generation; background editing and weather dynamics are mostly plausible, with some visible artifacts or limitations.
  - 81-100: excellent weather generation with natural background editing, convincing weather dynamics, coherent atmosphere, and high video quality.
  
  Return a JSON object:
  {
    "score": an integer from 0 to 100,
    "reason": "one concise explanation of the main strengths and weaknesses"
  }
  
  Scoring rule:
  0 = unusable or no meaningful weather generation
  100 = excellent weather generation with natural background editing,
        convincing weather dynamics, and high video quality
\end{promptverbatim}
\end{tcolorbox}
\vspace{-0.35em}
{\footnotesize\noindent\textbf{(4) VLM evaluation prompt.} The prompt scores weather background editing, weather dynamics, and video quality without rewarding unchanged backgrounds or measuring camera and object control.\par}

The main score $S_{\mathrm{VLM}}$ is the mean of Eq.~\ref{eq:vlm-evaluation}. The evaluation manifest may store textual reasons for failure analysis, but paper tables report only the scalar mean. The prompt discourages rewarding unchanged backgrounds because the Weather subset evaluates weather-state generation, not background copying.

\subsubsection{Human-Metric Alignment}
\label{app:human-metric-alignment}

To validate whether automatic weather metrics reflect human judgments, we compute the Pearson correlation coefficient ($r$) between human ratings and automatic scores on shared Weather-subset samples. Five annotators score each of the 100 Weather-subset samples on a five scales along weather correctness, weather dynamics, and geometry preservation under the generated weather, and we average the annotator scores before computing correlations. Table~\ref{tab-human-metric-alignment} reports the alignment for Weather Alignment and VLM Evaluation, where a higher correlation indicates better agreement with human assessments of weather visibility and overall weather-generation quality.

\begin{table}[H]
  \caption{\textbf{Human-metric alignment.} Pearson correlation coefficient ($r$) between automatic weather metrics and human ratings on the Weather subset.}
  \label{tab-human-metric-alignment}
  \centering
  \scriptsize
  \renewcommand{\arraystretch}{0.85}
  \setlength{\tabcolsep}{8pt}
  \begin{tabular}{lcc}
    \toprule
    Metric & Weather Alignment & VLM Evaluation \\
    \midrule
    Pearson ($r$) & 0.78 & 0.66 \\
    \bottomrule
  \end{tabular}
\end{table}

\subsubsection{User Study}
\label{app:human-preference}

Human preference complements the automatic metrics. Evaluators see the target weather prompt, optional input/source context, and two randomly ordered generated videos with hidden method names, then choose the video with more natural weather generation and weather-background editing. They are not asked to judge camera or object accuracy. For baseline $b\in\{\text{Cosmos-Transfer2.5},\text{Wan2.7-Edit}\}$, let $u_i^b\in\{0,1\}$ indicate that the majority vote prefers Holo-World over $b$ on sample $i$. We report
\begin{equation}
S_{\mathrm{User}}^b
=
\frac{100}{N}
\sum_{i=1}^{N}
u_i^b.
\label{eq:user-study}
\end{equation}
The main table keeps Holo-World vs. Cosmos-Transfer2.5 and Holo-World vs. Wan2.7-Edit separate.

\subsubsection{Background Preservation Metrics}
\label{app:background-preserve}

The Background columns in Tables~\ref{tab-core-ablation} and~\ref{tab-cfg-ablation} use rendered-background preservation metrics on the Real subset. These metrics evaluate source-supported background regions from rendered RGB controls; they are not Weather-subset scores.

For frame $t$, let $B_t\in[0,1]^{H\times W\times 3}$ denote the rendered reference background RGB frame and let $\hat{I}_t\in[0,1]^{H\times W\times 3}$ denote the generated frame. The renderer may contain invalid placeholder regions. We define an empty-color set $\mathcal{E}$ containing pixels close to the black placeholder and the gray placeholder used by the rendered controls. The valid background mask is the intersection of non-empty regions in the reference and generated frames:
\begin{equation}
M_t
=
\mathbf{1}\!\left[B_t\notin\mathcal{E}\right]
\land
\mathbf{1}\!\left[\hat{I}_t\notin\mathcal{E}\right].
\end{equation}
The same mask is applied to both images, so the metric reflects disagreement on comparable, source-supported background pixels rather than differences in unrendered regions.

LPIPS is sensitive to convolutional context around mask boundaries. We therefore composite both frames onto the same neutral color $c$ before computing image-level metrics:
\begin{align}
B_t^{M} &= M_t\odot B_t + (1-M_t)\odot c,\\
\hat{I}_t^{M} &= M_t\odot \hat{I}_t + (1-M_t)\odot c.
\end{align}
For the evaluated frame set $\mathcal{T}$, the reported background metrics are
\begin{align}
\mathrm{Background\mbox{-}PSNR}
&=
\frac{1}{|\mathcal{T}|}
\sum_{t\in\mathcal{T}}
\mathrm{PSNR}(B_t^{M},\hat{I}_t^{M}),\\
\mathrm{Background\mbox{-}SSIM}
&=
\frac{1}{|\mathcal{T}|}
\sum_{t\in\mathcal{T}}
\mathrm{SSIM}(B_t^{M},\hat{I}_t^{M}),\\
\mathrm{Background\mbox{-}LPIPS}
&=
\frac{1}{|\mathcal{T}|}
\sum_{t\in\mathcal{T}}
\mathrm{LPIPS}(B_t^{M},\hat{I}_t^{M}).
\end{align}
Higher Background PSNR/SSIM and lower Background LPIPS indicate better source-supported background preservation. The mask excludes unrendered regions and placeholder-color artifacts, making the metric better aligned with the Real-subset no-edit contract than whole-frame reconstruction.
\FloatBarrier